\documentclass{ieeeojies}

\newcommand{\ve}[1]{\mbox{\boldmath $ #1 $}}
\newcommand{\dve}[1]{\dot{\mbox{\boldmath $ #1 $}}}

\usepackage[pdftex]{graphicx}
\usepackage{subfigure}
\usepackage{cite}
\usepackage{amsmath,amssymb,amsfonts}
\usepackage{algorithmic}
\usepackage{textcomp}

\def\BibTeX{{\rm B\kern-.05em{\sc i\kern-.025em b}\kern-.08em
    T\kern-.1667em\lower.7ex\hbox{E}\kern-.125emX}}
\begin{document}

\title{Imitation Learning for Variable Speed Contact Motion for Operation up to Control Bandwidth}
\author{Sho \uppercase{Sakaino}\authorrefmark{1}, \IEEEmembership{Member, IEEE},
Kazuki \uppercase{Fujimoto}\authorrefmark{2}, \IEEEmembership{Non Member},
Yuki \uppercase{Saigusa}\authorrefmark{3}, \IEEEmembership{Non Member},
and Toshiaki \uppercase{Tsuji}\authorrefmark{2},
\IEEEmembership{Senior Member, IEEE}}
\address[1]{Faculty of Engineering, Information and Systems, Department of Intelligent Interaction Technologies, University of Tsukuba, Japan}
\address[2]{Electrical and Electronic Systems, Saitama University, Saitama, Japan}
\address[3]{Graduate School of Science and Technology, Degree Programs in Intelligent and Mechanical Interaction Systems, University of Tsukuba, Japan.}
\tfootnote{ This work was partly supported by the Adaptable and Seamless Technology Transfer Program through Target-driven R\&D (A-STEP) from the Japan Science and Technology Agency (JST) Grant Number JPMJTR20RG and the Japan Society for the Promotion of Science by a Grant-in-Aid for Scientific Research (B) under Grant 21H01347.}

\markboth
{Sakaino \headeretal: Preparation of papers for IEEE TRANSACTIONS and JOURNALS}
{Sakaino \headeretal: Preparation of papers for IEEE TRANSACTIONS and JOURNALS}

\corresp{Corresponding author: Sho Sakaino (e-mail: sakaino@iit.tsukuba.ac.jp).}

\begin{abstract}
The generation of robot motions in the real world is difficult by using conventional controllers alone and requires highly intelligent processing.
In this regard, learning-based motion generations are currently being investigated.
However, the main issue has been improvements of the adaptability to spatially varying environments, but a variation of the operating speed has not been investigated in detail.
In contact-rich tasks, it is especially important to be able to adjust the operating speed because a nonlinear relationship occurs between the operating speed and force (e.g., inertial and frictional forces), and it affects the results of the tasks.
Therefore, in this study, we propose a method for generating variable operating speeds while adapting to spatial perturbations in the environment.
The proposed method can be adapted to nonlinearities by utilizing a small amount of motion data.
We experimentally evaluated the proposed method by erasing a line using an eraser fixed to the tip of the robot as an example of a contact-rich task.
Furthermore, the proposed method enables a robot to perform a task faster than a human operator and is capable of operating close to the control bandwidth.
\end{abstract}


\titlepgskip=-15pt

\maketitle

\section{Introduction}\label{intro}
The utilization of machines and robots is promising; however, many processes are still performed manually, and labor is not yet fully automated because robots lack adequate environmental adaptability.
Particularly, contact-rich tasks, such as grinding and peg-in-hole, are difficult for robots.
Recent developments in reinforcement learning (RL) have succeeded in addressing the aforementioned tasks without precise prior knowledge of the tasks \cite{Zhang_grinding_2020, Lu_grinding_2019,Zhang_peg_in_hole_ra-l_2020, oikawa_2021_ra-l}.
However, this approach is impractical because the model learns with an enormous number of trials using actual machines. 
Given that robotic control involves interactions with the real-world environment, the time required for a single trial is constrained by the time constant of the physical phenomenon under investigation.
Hence, applying RL from the onset requires an impractical trial time.
If RL is performed in the simulation, this problem is greatly alleviated because there is no need to try it on actual machines.
Johannink \textit{et al.} showed that the number of trials required can be reduced using residual RL, and further reduced using sim2real, which transfers learning results from simulations to actual robots \cite{johannink_icra_2019}.
However, this still required several hundred to a thousand trials.

Imitation learning, which can address this problem, is gaining attention. 
In this process, humans provide demonstrations as teacher data, and the robots mimic human motion. This approach significantly reduces the number of trials required.
Many studies have demonstrated the effectiveness of imitation learning by applying Gaussian mixture models \cite{Calinon2007}\cite{kyrarini19:_robot_learn_indus_assem_task_human_demon}, neural networks (NNs) \cite{yang16:_repeat_foldin_task_human_robot}\cite{zhang18:_deep_imitat_learn_compl_manip}, and RL \cite{gupta2019relay}.
Some researchers have reported visual imitation learning \cite{jiny20:_geomet_persp_visual_imitat_learn}.
Imitation learning using force information has also attracted notable attention owing to its high adaptability to environmental changes \cite{lee15:_learn_force_based_manip_defor, kormushev11:_imitat_learn_posit_force_skill,ochi18:_deep_learn_scoop_motion_using_bilat_teleop,rozo19:_inter_trajec_adapt_force_baysian_optim, osa18:_onlin_trajec_plann_force_contr}.
It is worth noting that it may be possible to reduce the number of required motion teachings if the policy is learned in advance using simulation-based methods such as \cite{shahid_smc_2020} before imitation learning.
However, for its implementation, it is necessary to create a robot and environment simulator.
Particularly, it is difficult to create an accurate simulator when there are flexible objects in the environment.

The above-mentioned imitation learning is focused on performing geometrically challenging robotic tasks and is not relevant to reproducibility over time, such as in the case of a phase delay.
Consequently, the movements are often static and slower than human operations, and it is difficult to realize movements based on the dynamic interaction between robots and objects.
Motion that considers friction and inertial forces, such as that described in \cite{Tsuji2016}, remains a challenge.
Conventional imitation learning predicts the next response value of a robot and provides it as a command value. 
Generally, no ideal control system exists, and a delay between the command and response values occurs. 
Consequently, only low-speed operation, wherein control systems can be assumed to be ideal, can be achieved.

We recently showed that this problem can be solved using four-channel bilateral control \cite{adachi18:_imitat_learn_objec_manip_based}\cite{fujimoto19:_time_series_motion_gener_consid}.
Bilateral control is a remote operation that synchronizes two robots: a master and a slave.
Furthermore, four-channel bilateral control is a structure with position and force controllers implemented on both robots \cite{sakaino11:_multi_dof_micro_macro_bilat} \cite{sakaino17:_bilat_contr_elect_hydraul_actuat}.
Using bilateral control, an operator can experience a control delay on the slave side and dynamic interaction with the environment. 
Thus, the operator can compensate for the control delay and dynamic interaction.
There are conventional methods of imitation learning using bilateral control \cite{rozo13}\cite{ochi18:_deep_learn_scoop_motion_using_bilat_teleop}.
However, even if bilateral control is used, it is inadequate.
We revealed that the teacher data obtained via bilateral control can be fully utilized under the following three important conditions:
\begin{enumerate}
 \item Predicting the master robot's response\\
The command in the next step must be predicted when the response of a certain slave is measured.
In the case of bilateral control, the response value of a master is given as the command value of the slave, and the command value can be directly measured. 
It should be noted that this command value includes human skills to compensate for control delays and dynamic interactions.
 \item Having both position and force control in a slave\\
position control is robust against force perturbations, and force control is robust against position perturbations.
Although robot control can be described as a combination of these controls \cite{sakaino13:_novel_motion_equat_gener_task}, the predominant control is task-dependent and often not obvious.
In this case, machine learning must apply a configuration that can adjust to both position and force commands.
 \item Maintaining control gains\\
Research has also been conducted on adjusting control gains to achieve environmental adaptability \cite{rozo15:_learn}.
However, if the control gains are changed, the dynamic characteristics of the control also change.
Robots are then unable to mimic the skills of humans and compensate for control delays and dynamic interactions.
In summary, the controllers must be consistently applied when the training data are collected and during autonomous execution. 
\end{enumerate}
Our method satisfies these requirements, and the control system does not need to be ideal because the operation is performed by explicitly considering the control delay by predicting the response of the master.
Therefore, it is possible to realize object operation at a rate comparable to that of humans, and high adaptability to environmental changes is achieved.
A detailed explanation can be found in \cite{ayumu20:_imitat_learn_based_bilat_contr}\cite{sasagawa21:_motion_gener_using_bilat_contr}.

Given that fast motion can be achieved using the proposed method, a generalization ability with respect to the operating speed is the next target.
A basic study on achieving variable operating speed was proposed by Yokokura \textit{et al.} \cite{yokokura08:_motion_copyin_system_real_haptic_variab_speed}, in which a robot moved autonomously by reproducing stored motion. 
Reproduced motion was generated using simple linear interpolation and extrapolation of the stored motion. However, this method has been evaluated only in highly transparent, single-degree-of-freedom (DOF) linear motors. 
In actual multi-DOF robots, dynamic forces, such as the inertial force, change significantly according to the operating speed. 
For example, in a line-erasing task, the required state of the end-effector differs depending on the operating speed because the pressing force on the paper surface is adjusted to utilize inertial force during high-speed operation, and the eraser is actively pressed against the paper surface during low-speed operation.
The force and operating speed clearly have a nonlinear relationship. 
However, it should be possible to express this relationship using specific functions.
Percio \textit{et al.} also demonstrated variable speed operation \cite{Percio_iros_2020}; however, this method did not effectively reproduce the operation speed. Consequently, it was not possible to operate at the frequency of the control bandwidth.
The self-organization of the operation speed is realized by parametric biases, in which the physical parameters of robotic motions can be adjusted \cite{tani03:_self_organ_behav_primit_multip_attrac_dynam}.
However, because this is a self-organizing process, it is not possible to precisely command the desired operating speed.

In this study, we propose a method in which the operating speed is varied using imitation learning based on four-channel bilateral control. 
It should be noted that in the proposed method, the operating speed can exceed that of the original demonstrations and even be capable of operating at the frequency of the control bandwidth.
If a robot can move quickly, its productivity can be improved. Moreover, it is also desirable to adjust the operating speed to match the production speed of other production lines.
To evaluate the effectiveness of the proposed method, we performed a task in which a robot erased a line written in pencil using an eraser fixed to the robot as an example of contact-rich motions.
Using the relationship between the inertial force, friction force, and operating speed is necessary to accomplish this task because a large operating force is required for fast operation to compensate for the inertial force and vice versa.
Moreover, when the same task is performed, a different operating force is required depending on the operating speed because the friction characteristics change significantly with speed \cite{ruderman_TIE_2016}\cite{mostefai_TMECH_2010}.
Using the proposed method, the operating speed is determined based on the peak frequency calculated using the fast Fourier transform (FFT), and the slave responses are concatenated and inputted into an NN. 
Variable speed operation is achieved by incorporating the operating frequency as an input.
This is a further development of our imitation learning using bilateral control.

It is worth noting that it is not difficult to achieve variable-speed contact control with known dynamics or to operate faster than humans; however, it is difficult to operate up to the control bandwidth.
Additionally, it is not difficult to slowly control contact with unknown dynamics.
Therefore, the contributions of this study are summarized as follows:
\begin{itemize}
 \item Variable speed contact motion with unknown dynamics
 \item Contact motion with unknown dynamics up to the control bandwidth
\end{itemize}

It is also worth noting that many previous studies on object manipulation using machine learning have mainly aimed at improving the generalization performance for space, and few of them have aimed at improving the generalization performance over time.

The remainder of this paper is organized as follows. 
Section 2 presents the robot control system and bilateral control used in this study. 
Section 3 presents the proposed learning method and detailed network structure.
Section 4 details the experiment and results in addition to a description of a comparative experiment involving the proposed method and a variable-speed motion copy approach based on the study described in \cite{yokokura08:_motion_copyin_system_real_haptic_variab_speed}.
Finally, Section 5 presents the concluding remarks and areas of future study.

\section{Robot and controller} \label{1:setup}
In this section, the robots and controllers used in this study are presented.
\subsection{Setup} \label{2.1:manipulator}
\begin{figure}[t] 
  \centering 
  \includegraphics[width=70mm]{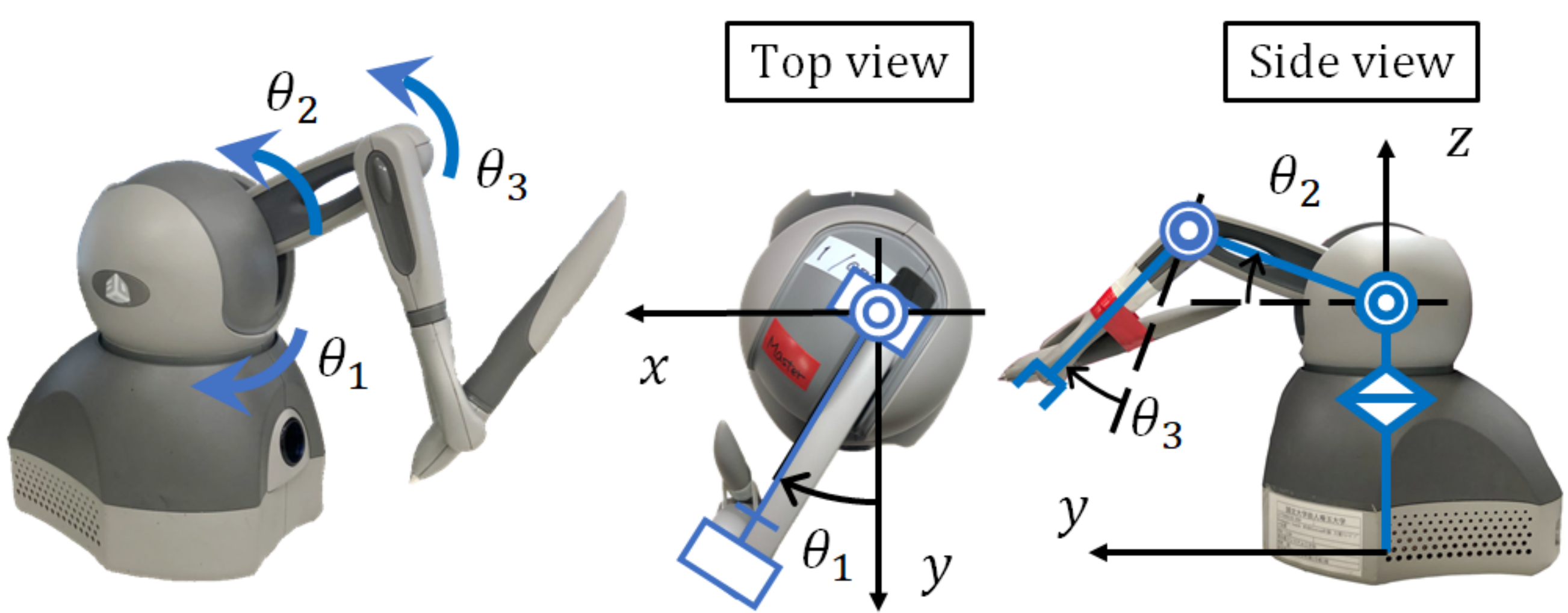}
  \caption{Definition of the robot's joints and Cartesian coordinates}
  \label{define}
\end{figure}
In this study, we used two Geomagic Touch haptic devices manufactured by 3D systems (Rockhill, SC, USA) as manipulators (Fig.~\ref{define}, respectively. 
Sensors, a motor driver, and a microcomputer were built into the robot, and the robot could be controlled by connecting a USB cable to a personal computer.
Detailed specifications can be found at https://www.3dsystems.com/haptics-devices/touch/specifications.
Two robots were used during the data collection phase, and an autonomous operation phase using an NN model was executed using a single robot. 
The robot's joints and Cartesian coordinates are defined as shown in Fig.~\ref{define}.
The model of the robots was assumed to be the same as that in \cite{sasagawa21:_motion_gener_using_bilat_contr}.
However, the physical parameters of the robot were different and were identified on the basis of \cite{yamazaki17:_estim_kinet_model_human_arm}.
Note that in this model, we neglected the interference among the axes and treated them as three single-DOF axes.
Therefore, inertia, damping, and gravity coefficients were identified as a single-DOF system.
\begin{table}
\centering
\caption{Identified system parameters}
\label{tbl:id_parameter}
\begin{tabular}{@{}l|cccccc}
\hline
      $J_{1}$ & Joint~1 inertia [{\rm m Nm}] & 3.49 \\
      $J_{2}$ & Joint~2 inertia [{\rm m Nm}] & 3.36 \\
      $J_{3}$ & Joint~3 inertia [{\rm m Nm}] & 1.06 \\
      $D$ & friction compensation coefficient [\rm m kg m$^2$/s] & 12.1 \\
      $G_{1}$ & Gravity compensation coefficient~1 [{\rm m Nm}] & 124 \\
      $G_{2}$ & Gravity compensation coefficient~2 [{\rm m Nm}] & 51.6 \\
      $G_{3}$ & Gravity compensation coefficient~3 [{\rm m Nm}] & 81.6 \\
\hline
\end{tabular}
\end{table}
Table~\ref{tbl:id_parameter} lists the physical parameter values used in this study.
The parameters $J$, $D$, and $G$ are the inertia, friction compensation coefficient, and gravity compensation coefficient, respectively.
The parameters with subscripts 1, 2, and 3 represent those of the first, second, and third joints, respectively.

\subsection{Controller}
\begin{figure}[t] 
  \centering 
  \includegraphics[width=90mm]{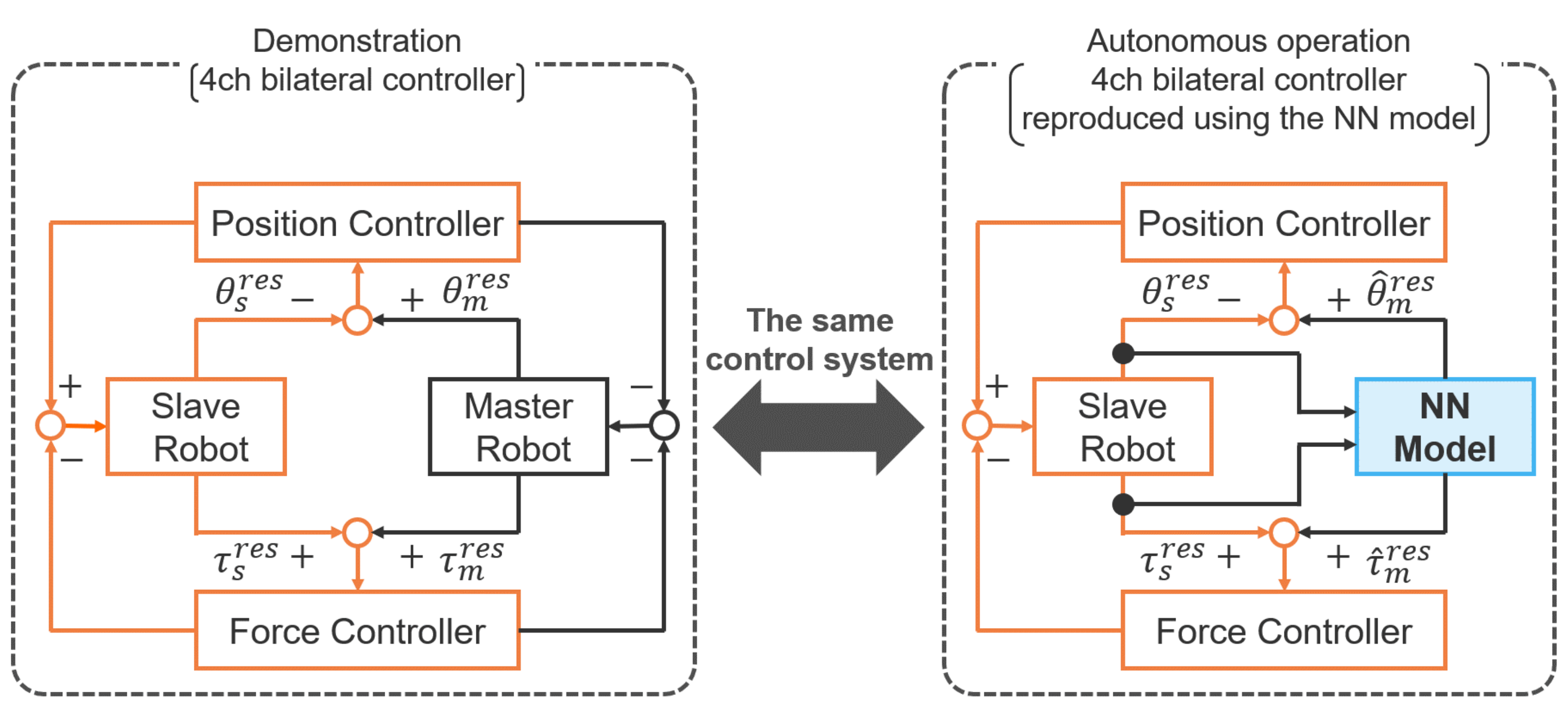}
  \caption{
Four-channel bilateral controller.
The figure on the left shows a four-channel bilateral controller that was used in the demonstrations.
The figure on the right represents a situation of autonomous operation.
In autonomous operations, the master robot and master controllers are substituted by an NN model to mimic the master responses.
It should be noted that the systems including the slave robot and slave controllers
(orange lines) are the same for both figures.
}
  \label{4ch}
\end{figure}
This robot can measure the joint angles of the first to third joints and calculate the angular velocity and torque response using pseudo-differentiation and a reaction force observer (RFOB) \cite{RTOB}, respectively. 
Acceleration control was realized using a disturbance observer (DOB) \cite{DOB}, and robustness against modeling errors was guaranteed.
A position controller and a force controller were implemented in the robot; these two controllers were composed of a proportional and differential position controller and a proportional force controller, respectively.
Here, $\theta$, $\dot{\theta}$, and $\tau$ represent the joint angle, angular velocity, and torque, respectively, and the superscripts $cmd$, $res$, and $ref$ indicate the command, response, and reference values, respectively.
The torque reference of the slave controller $\ve{\tau}_s^{ref}$ is given as follows:
\begin{eqnarray}
\label{eq:slave_controller}
\ve{\tau}_s^{ref}=  \ve{J}(K_p+K_d s)(\ve{\theta}_s^{cmd}-\ve{\theta}_s^{res})+K_f(\ve{\tau}^{cmd}_s-\ve{\tau}^{res}_s),
\end{eqnarray}
where $\ve{\theta}_s$ and $\ve{\tau}_s$ are the slave variables, defined as follows:
\begin{eqnarray}
 \ve{\theta_s}=
\left[
  \begin{array}{c}
  \theta_{s1} \\
  \theta_{s2} \\
  \theta_{s3} \\
  \end{array}
  \right]
,
 \ve{\tau_s}=
\left[
  \begin{array}{c}
  \tau_{s1} \\
  \tau_{s2} \\
  \tau_{s3} \\
  \end{array}
  \right].
\end{eqnarray}
In addition, $\ve{J} = \textrm{diag} [{J_1, J_2, J_3}]$.
Here, $s$ is the Laplace operator.
Additionally, the parameters $K_p$, $K_d$, and $K_f$ are scalar, and represent the proportional position gain, derivative position gain, and proportional force gain, respectively.
We set the same control parameters in each axis.
The gains were determined by trial and error so that the control bandwidth covered the motion frequencies of a human.
Note that	we adjusted the proportional-derivative control of the position to have the characteristics of critical damping.
Table~\ref{tbl:id_parameter} shows the control parameters.

\begin{table}
\centering
\caption{Gain of robot controller}
\label{tbl:gain}
\begin{tabular}{c|ccc}
 \hline
      $K_{p}$ & Position feedback gain & 121 \\
      $K_{d}$ & Velocity feedback gain & 22.0 \\
      $K_{f}$ & Force feedback gain & 1.00 \\
      $g$ & Cut-off frequency of pseudo derivative [{\rm rad/s}] & 40.0 \\ 
      $g_{DOB}$ & Cut-off frequency of DOB [{\rm rad/s}] & 40.0 \\ 
      $g_{RFOB}$ & Cut-off frequency of RFOB [{\rm rad/s}] & 40.0 \\ \hline
\end{tabular}
\end{table}
Bilateral control is a remote operation technology between two robots. 
The operator first operates the master robot and then operates the slave robot directly through the master robot \cite{sakaino11:_multi_dof_micro_macro_bilat} \cite{sakaino17:_bilat_contr_elect_hydraul_actuat}.
The operation and reaction forces can be independently measured by the master and slave.
This controller was implemented to imitate human object manipulation skills. 
A four-channel bilateral controller was implemented similar to that in \cite{sasagawa21:_motion_gener_using_bilat_contr}.

A block diagram of the four-channel bilateral controller in the demonstration (data collection phase) is shown on the left side of Fig.~\ref{4ch}.
The command values of the slave robot in the four-channel bilateral control are given as follows:
\begin{eqnarray}
 \ve{\theta}^{cmd}_s=\ve{\theta}_m^{res},
 \ve{\tau}^{cmd}_s=-\ve{\tau}_m^{res},
\end{eqnarray}
where $\ve{\theta}_m$ and $\ve{\tau}_m$ are the master variables defined as follows:
\begin{eqnarray}
 \ve{\theta_m}=
\left[
  \begin{array}{c}
  \theta_{m1} \\
  \theta_{m2} \\
  \theta_{m3} \\
  \end{array}
  \right],
 \ve{\tau_m}=
\left[
  \begin{array}{c}
  \tau_{m1} \\
  \tau_{m2} \\
  \tau_{m3} \\
  \end{array}
  \right].
\end{eqnarray}

\section{Imitation Learning for Variable-Speed Operation} \label{3:proposed_method}
We propose a method for generating variable-speed motion that can exceed the speed of the original motion using imitation learning.

\begin{figure}[t] 
  \centering 
  \includegraphics[width=40mm]{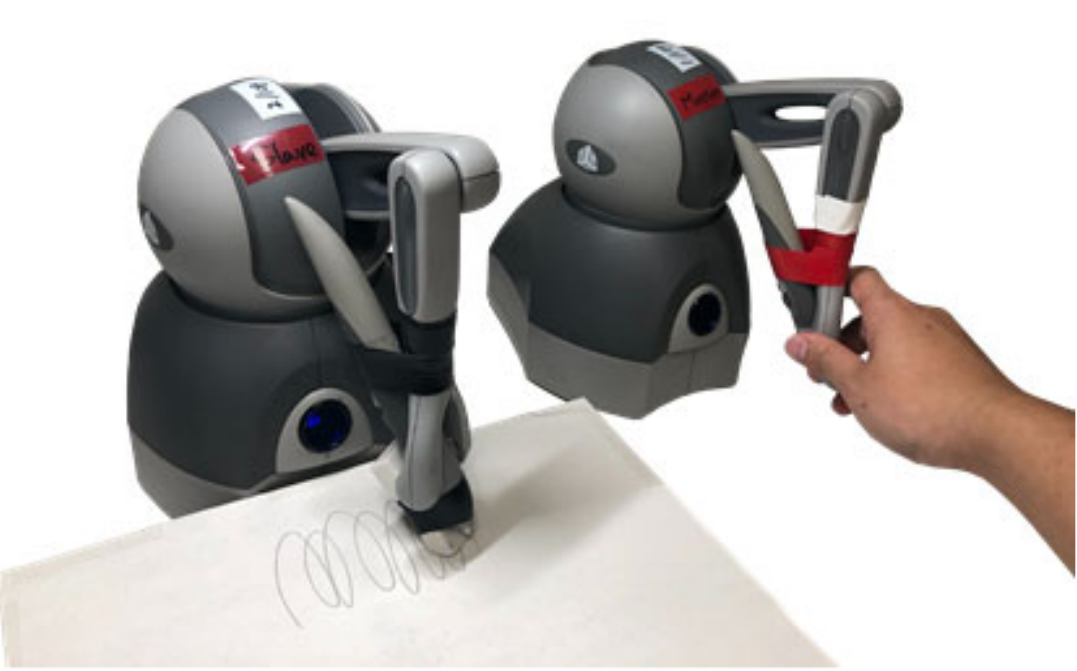}
  \caption{Data collection using four-channel bilateral control}
  \label{bilate_photo}
\end{figure}
\begin{figure*}[t] 
  \centering 
  \includegraphics[width=17cm]{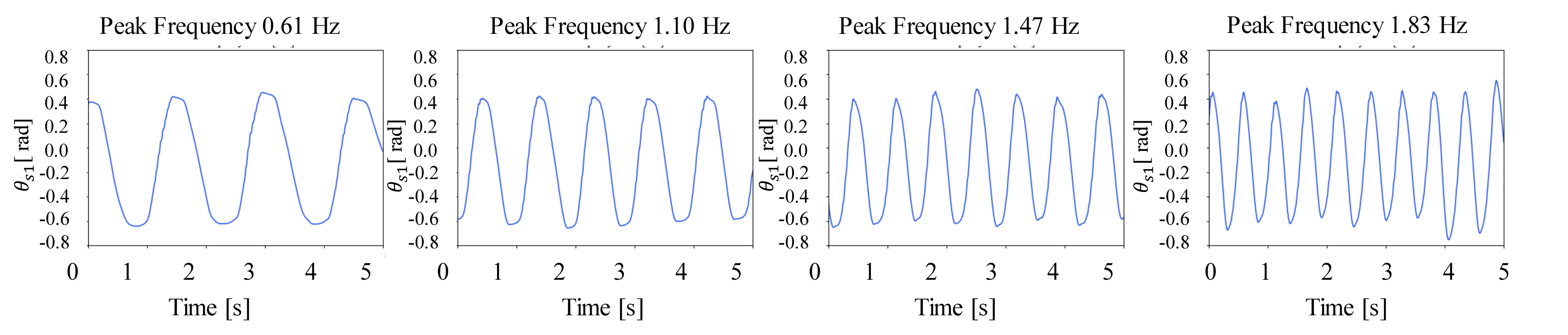}
  \caption{$\theta_{s1}^{res}$ in training data at each frequency for a height of 5.6 cm. The amplitude of the angular response was almost identical.}
  \label{ang_BPM}
\end{figure*}
\begin{figure*}[t] 
  \centering 
  \includegraphics[width=17cm]{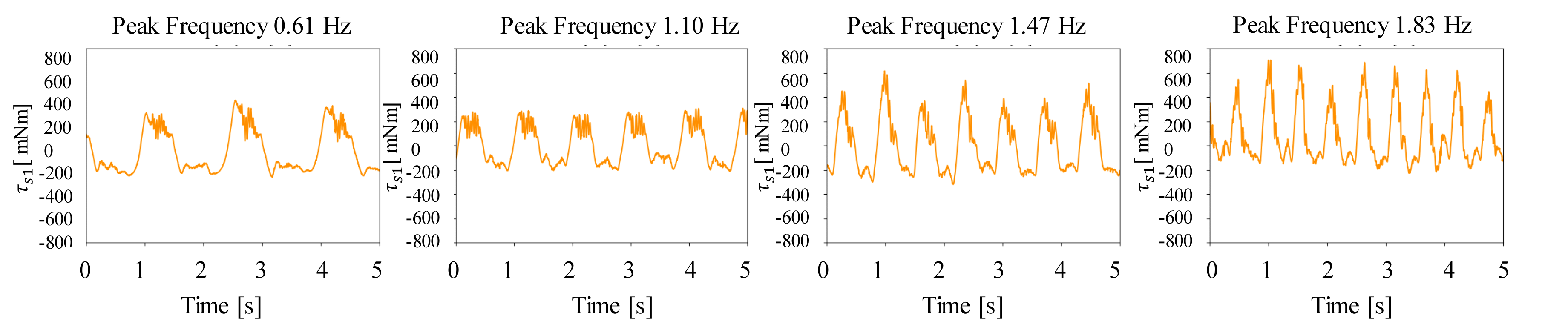}
  \caption{$\tau_{s1}^{res}$ in training data at each frequency for a height of 5.6 cm. The amplitude of the torque response varies according to the frequency.}
  \label{for_BPM}
\end{figure*}


\subsection{Data collection} \label{3.1:collection}
Fig.~\ref{bilate_photo} shows the data collection phase.
The two robots were used for data collection based on a four-channel bilateral control, as described in Section \ref{2.1:manipulator}. 
The objective was to generate motion to quickly or slowly erase a line written in pencil. 
This task simulated contact-rich motions, such as polishing \cite{Li_ieeeaccess_2020}, grinding \cite{Li_cybernetics_2016}, and wiping \cite{Dometios_RA-L_2018}.
Therefore, the operator of the master robot erased the lines using seven different frequencies, that is, 0.61, 0.85, 1.10, 1.22, 1.47, 1.59, and 1.83 Hz. Frequency adjustments were performed using a metronome.
The motions at 0.61 and 1.83 Hz were the slowest and fastest motions that the human was able to manipulate the robot, respectively, and for the other frequencies, the human was instructed to divide these frequencies into seven equal parts using a metronome. 
It was not exactly divided into seven equal parts, as the value was calculated based on the peak frequency of human motion.
This trial was conducted three times at paper heights of 3.5, 5.6, and 6.6 cm from the surface of the desk.
The height was adjusted by piling up books under a plate for placing paper.
A total of 21 trials were conducted.
The saved motion data points were acquired over 15 s in each case, and the joint angle, angular velocity, and torque of the master and slave data were stored at 1 kHz. 
Training data were obtained by augmenting the collected data 20 times by down-sampling at 50 Hz using the technique described in \cite{rahmatizadeh_2018}.

Additionally, Figs. \ref{ang_BPM} and \ref{for_BPM} show some of the training data of $\theta_{s1}^{res}$ and $\tau_{s1}^{res}$ for a height of 5.6 cm, respectively. 
From these figures, when the operating speed changes, it can be confirmed that the required motion and force adjustment differ, although the trajectory is similar.
When the operation is the fastest, the torque is the greatest because the inertial force is the highest, whereas the torque decreases with a decrease in the operating frequency.
However, when the operation was the slowest, the torque was slightly larger to compensate for the nonlinearity of the frictional force.
This is a major problem, which makes it difficult to achieve motion generation at variable speeds.
\subsection{Training the NN model phase} \label{3.2:learning}
In this study, we use a network consisting of a recurrent NN (RNN). 
An RNN, which has a recursive structure, is a network that holds time-series information. 
This network has contributed significantly to the fields of natural language processing and speech processing \cite{sundermeyer15:_from_feedf_recur_lstm_neural}\cite{zhang16:_highw_rnns} and has recently been widely applied to robot motion planning \cite{zhang17:_three_recur_neural_networ_three}.
However, RNNs are hindered by the vanishing gradient problem, which makes it difficult to learn long-term data.
Long short-term memory (LSTM) refers to an NN that can learn long-term inferences \cite{hochreiter97:_long_short_term_memor}. This approach was improved based on the results of numerous studies and was adopted in this study.
To extract the feature values from the response variables that do not depend on time-series information, we implemented a convolutional NN (CNN) prior to the LSTM.
We expected that the CNN would extract time-independent transformations, such as Jacobian matrices.

The network inputs are $\ve{\theta}_s^{res}$, $\dve{\theta}_s^{res}$, $\ve{\tau}_s^{res}$, and the frequency command of the first joint, whereas the outputs are $\hat{\ve{\theta}}_m^{res}$, $\hat{\dve{\theta}}_m^{res}$, and $\hat{\ve{\tau}}_m^{res}$ of each joint in the next step.
The variables with $\hat{\bigcirc}$ are estimates given by the NN.
The frequency command was designed based on the peak frequency values of the first joint angle of the robot, which was calculated using the FFT. 
Here, the next step indicates a point 20 ms later than the slave data.
Autonomous operation was realized by considering the network calculation time required to generate online motion.
Thus, the data had 315,000 ($15,000 [\rm{ms}] / 20 [\rm{ms}] \times 21 [\rm{trials}] \times 20 [\rm{augmentation}]$) input-output samples.
Additionally, the weights were optimized using the mean square error between the normalized master value and network output.

\begin{figure*}[t] 
  \centering 
  \includegraphics[width=17cm]{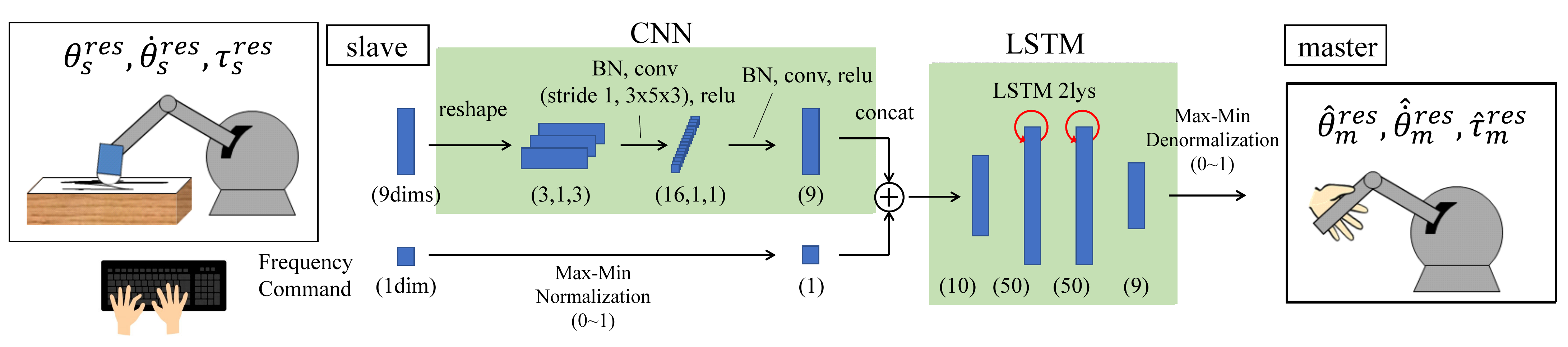}
  \caption{Overview of proposed NN using CNN for normalization}
  \label{cnn_model}
\end{figure*}
Fig.~\ref{cnn_model} shows the network.
The responses $\ve{\theta}^{res}_s$, $\dve{\theta}_s^{res}$, and $\ve{\tau}_s^{res}$ of each joint of the slave robot are reshaped into other channels.
The reshape was designed to predict the effect of batch normalization (BN) for each unit dimension.
Additionally, the mini-batch consisted of 100 random sets of 300 time-sequential samples corresponding to 6 s. 
The frequency command was manually provided using a keyboard and normalized using max-min normalization.
Max-min denormalization was set at the output of the network.
In this study, the computer used for training and autonomous operation comprised an Intel Core i7-8700K CPU, 32 GB of memory, and an nVIDIA GTX 1080 Ti GPU.
\begin{figure}[t]
  \centering 
  \includegraphics[width=7cm]{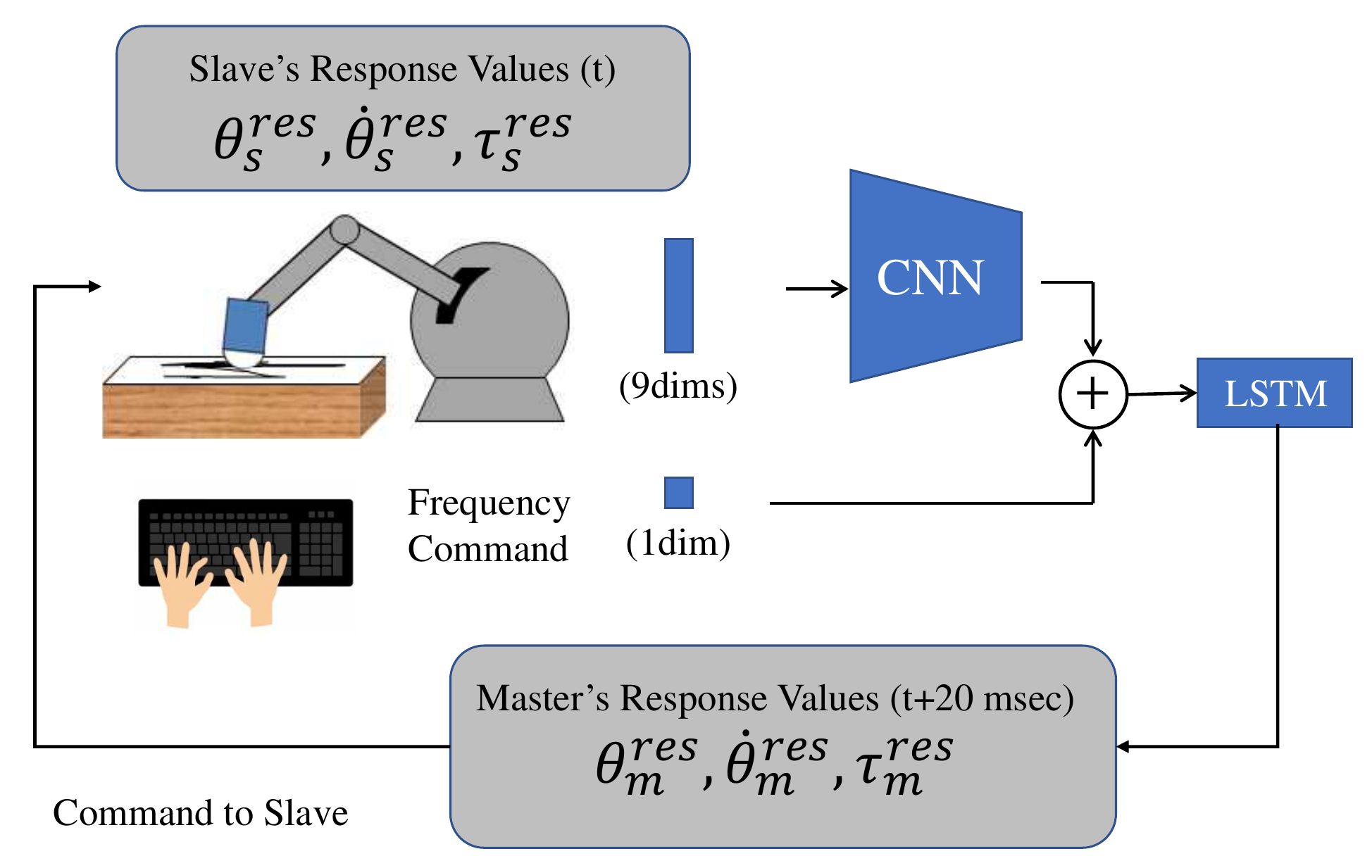}
  \caption{Autonomous operation phase}
  \label{autonomous}
\end{figure}

\subsection{Autonomous operation phase} \label{3.3:execution}
The right part of Fig.~\ref{4ch} shows a block diagram of the slave robot conducting autonomous execution using the trained NN. 
In the autonomous operation phase, the demonstrator, master robot, and master controllers are substituted by the trained NN.
In this case, the command values are not the true response values of the master, but the estimated values provided by the NN.
Fig.~\ref{autonomous} shows its detailed schematic.
The NNs predict responses of the master in the next step (20 ms after) from the responses of the slave and frequency commands.
Then, the master's responses are commanded to the slave.

\begin{eqnarray}
 \ve{\theta}^{cmd}_s=\hat{\ve{\theta}}_m^{res},
 \ve{\tau}^{cmd}_s=-\hat{\ve{\tau}}_m^{res}
\end{eqnarray}
It is worth noting that the control system in the autonomous execution phase is the same as that in the data collection phase.
Although the input interval into the network was 20 ms, the control period of the position and force controller was 1 ms.
Therefore, the control input was updated during every control period.

\section{Experiment} \label{4:experiment}
This section presents the experimental evaluation of the proposed method.

\subsection{Control bandwidth}
\begin{figure}[tb]
  \centering 
  \includegraphics[width=7cm]{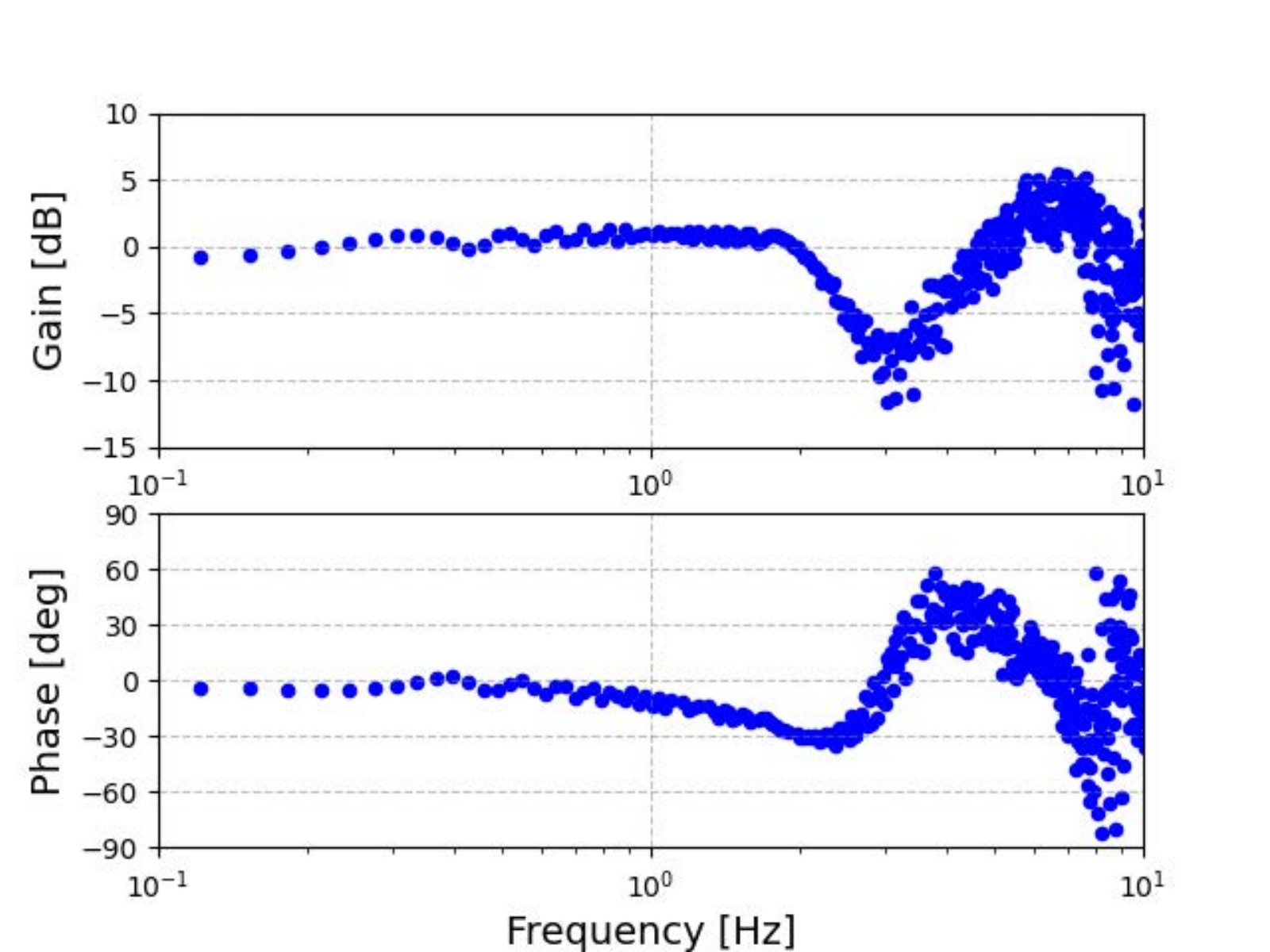}
  \caption{Bode diagram of $\theta_{s1}^{res} / \theta_{s1}^{cmd}$ of free motion}
  \label{bode_diagram}
\end{figure}

First, the control bandwidth of the robot was tested.
Fig.~\ref{bode_diagram} shows a Bode diagram of the angle response of the first joint ($\theta_{s1}^{res} / \theta_{s1}^{cmd}$) in free motion with zero torque commands.
As can be observed from this figure, the control bandwidth where the gain is -3 dB was approximately 2.3 Hz.
It should be noted that operations over the control bandwidth are extremely difficult, and motion over the control bandwidth is not a subject of interest in the research field of control engineering.

\subsection{Preliminary experiment}
\begin{figure}[tb]
  \centering 
  \includegraphics[width=7cm]{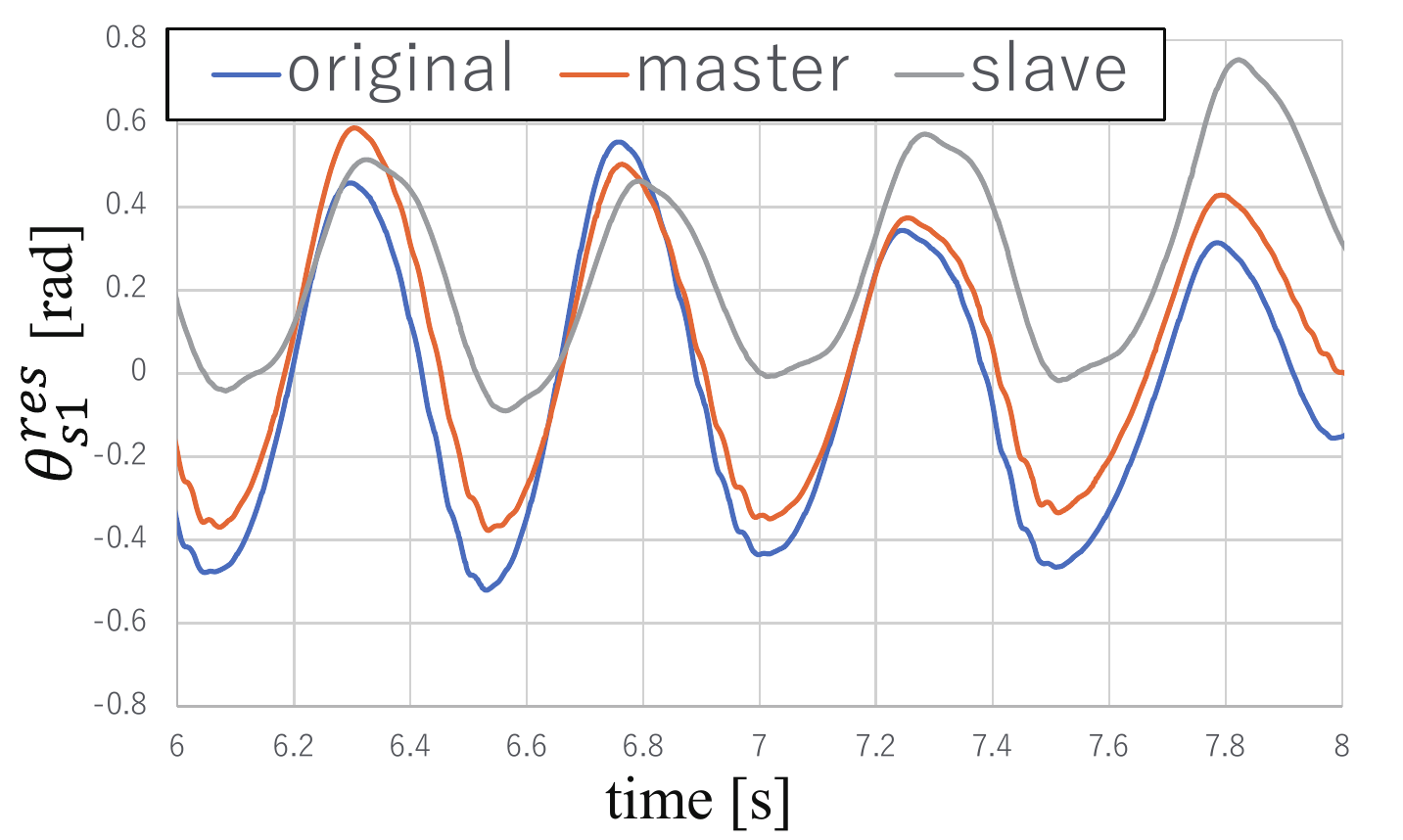}
  \caption{Result of $\theta_{s1}^{res}$ of preliminary experiment. A fast motion could be achieved if and only if we predicted the master values as commands.}
  \label{motion_copy_theta}
\end{figure}

\begin{figure}[tb]
  \centering 
  \includegraphics[width=7cm]{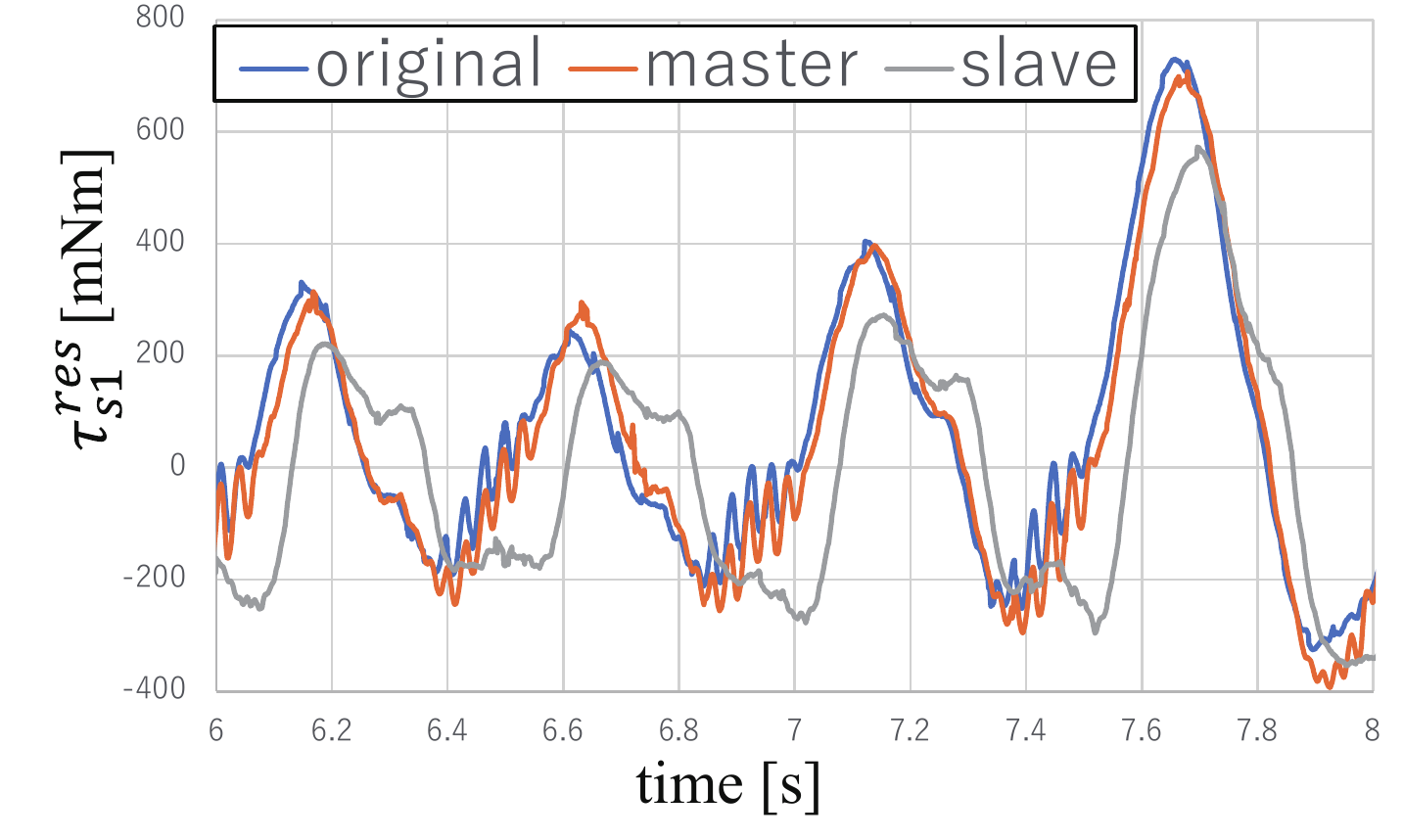}
  \caption{Result of $\tau_{s1}^{res}$ of preliminary experiment. A fast motion could be achieved if and only if we predicted the master values as commands.}
  \label{motion_copy_tau}
\end{figure}

Prior to performing the main experiment, one of the three conditions described in the introduction was tested.
The validity of the predicting command value of the slave, that is, the predicting master response, was evaluated.
In this case, the environment was not changed from the data collection phase, the NN was not used, and motion at 2.17 Hz was simply replayed as a command value.
The following two command values for the next step were given for comparison:
\begin{enumerate}
 \item the master's response during the training data (our approach); $\ve{\theta}^{cmd}_s=\ve{\theta}_m^{tr},
 \ve{\tau}^{cmd}_s=-\ve{\tau}_m^{tr}$, and
 \item the slave's response during the training data; $\ve{\theta}^{cmd}_s=\ve{\theta}_s^{tr},
 \ve{\tau}^{cmd}_s=\ve{\tau}_s^{tr}$.
\end{enumerate}
The variables with superscript $tr$ indicate the training data.
Figs.~\ref{motion_copy_theta} and \ref{motion_copy_tau} show the experimental results for $\theta_{s1}^{res}$ and $\tau_{s1}^{res}$, where the blue lines represent the original slave responses in the training data.
The orange and gray lines represent the responses reproduced using the master and slave responses in the training data, respectively.
The orange lines indicate that the response was almost identical to that of the data collection when the master's next response was used as a command value.
However, based on the gray lines, when the next response of the slave was given as a command value, the shape of the response differed significantly from that in the training data.
It is evident that the amplitude was smaller than that in the original slave response, and a large phase delay occurred.
Given that the motion was rapid, and it was very close to the control bandwidth, the transfer function from the command to the response cannot be 1.

These results clearly show that predicting the master's response is important for reproducing fast motion.
It should be noted that kinesthetic teaching cannot satisfy this condition, nor does conventional imitation learning using bilateral control \cite{rozo13}.
As such, temporal reproducibility at high speeds can only be achieved using our approach.
Hence, variable-speed imitation learning with precise reproducibility has been made possible for the first time.


\subsection{Comparative experiment} \label{4.2:ex1}
The results of the experiment conducted to change the operating speed based on the training data were compared with the results of a motion copying system \cite{yokokura08:_motion_copyin_system_real_haptic_variab_speed}.
In the latter, the data collected at a frequency of 1.22 Hz and height of 5.6 cm were used to reproduce the operation.
Given that the motion copying system simply rescales the time axis, it only requires one time series of data for reproduction. 
To convert the operating speed, the original data were rescaled to fit the target speed data.
The training data were rescaled along the time axis of the data using linear interpolation with a zero-order hold.

\begin{figure}[t]
  \centering 
  \includegraphics[width=6cm]{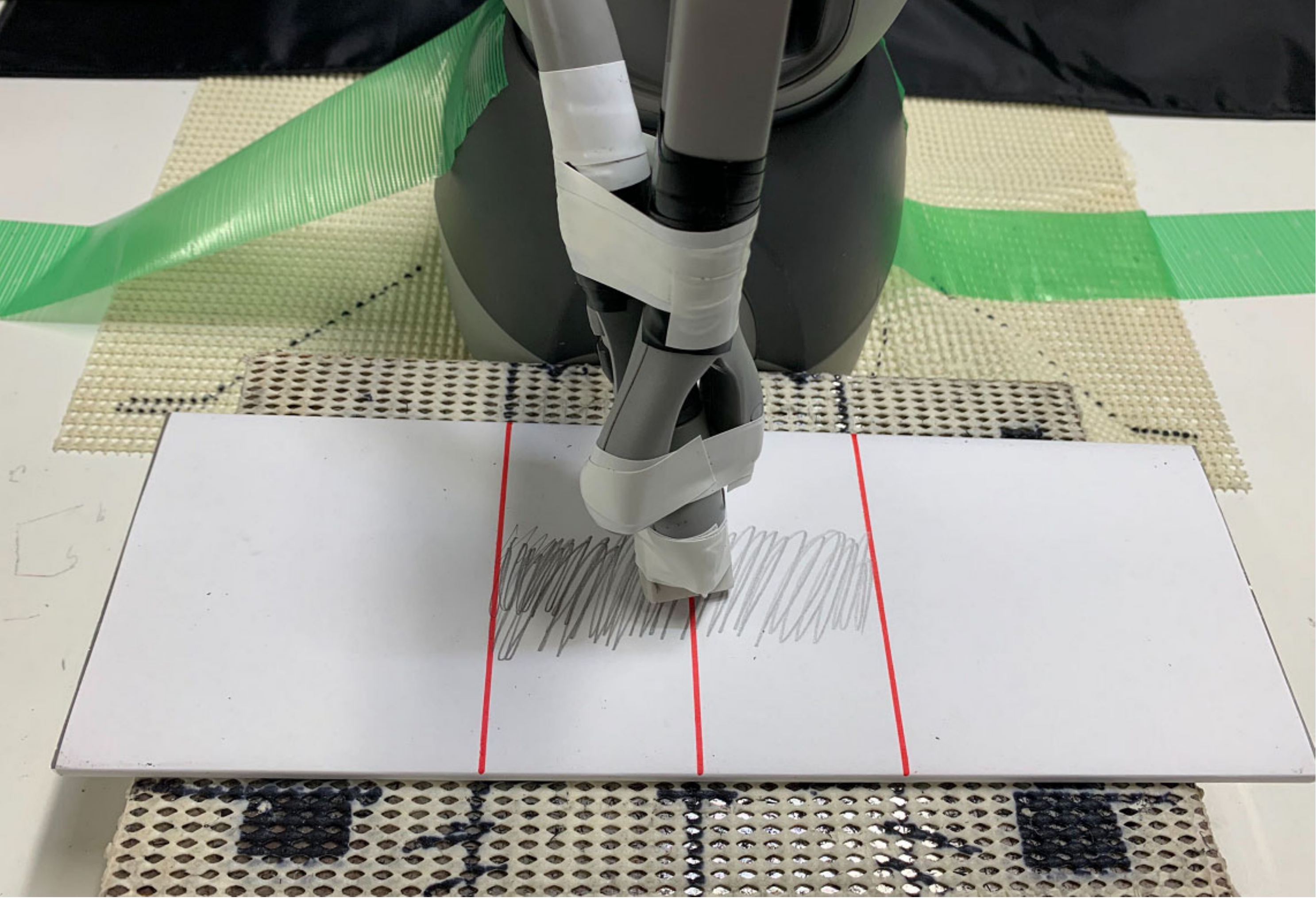}
  \caption{Working area. When the robot erased more than 90 \% of the area inside the red lines, it was considered a success. Note that we do not intend to erase all black lines. However, we investigated whether the robot could erase the arc-shaped area.}
  \label{area}
\end{figure}

\begin{figure}[t]
  \centering 
  \includegraphics[width=7cm]{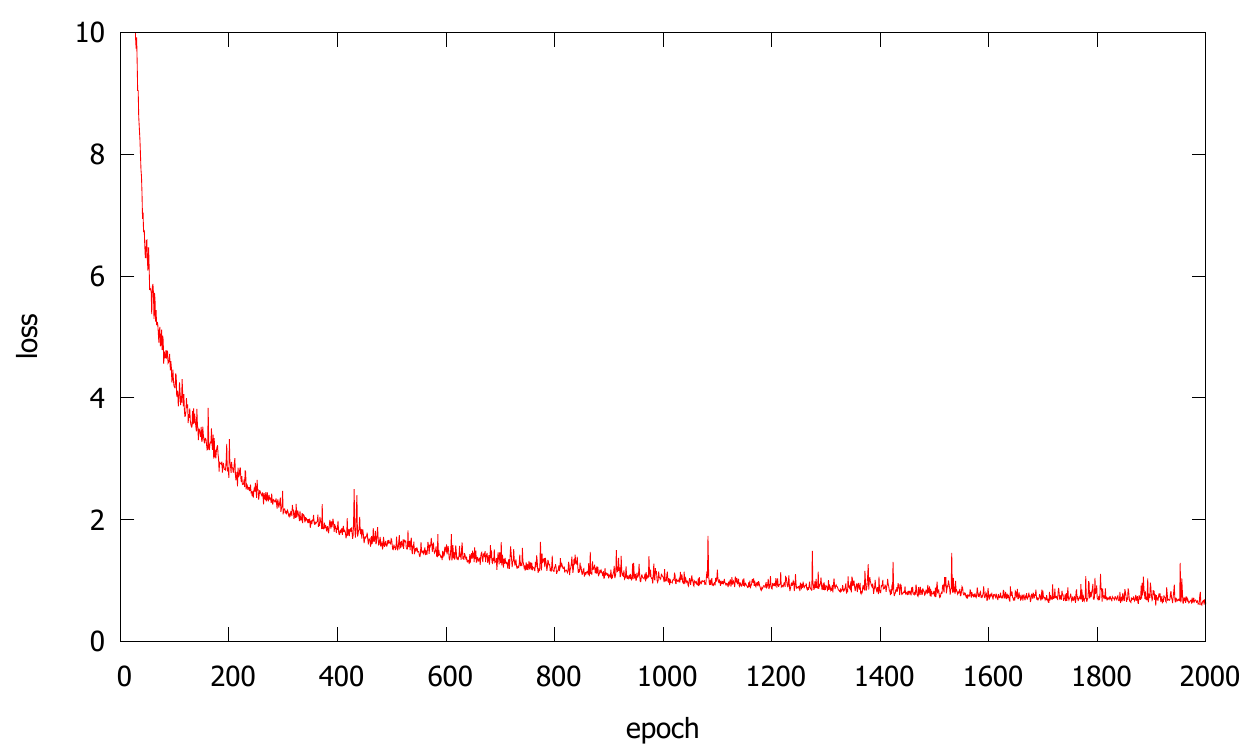}
  \caption{Training loss}
  \label{loss}
\end{figure}

\begin{figure*}[t]
  \centering 
  \includegraphics[width=18cm]{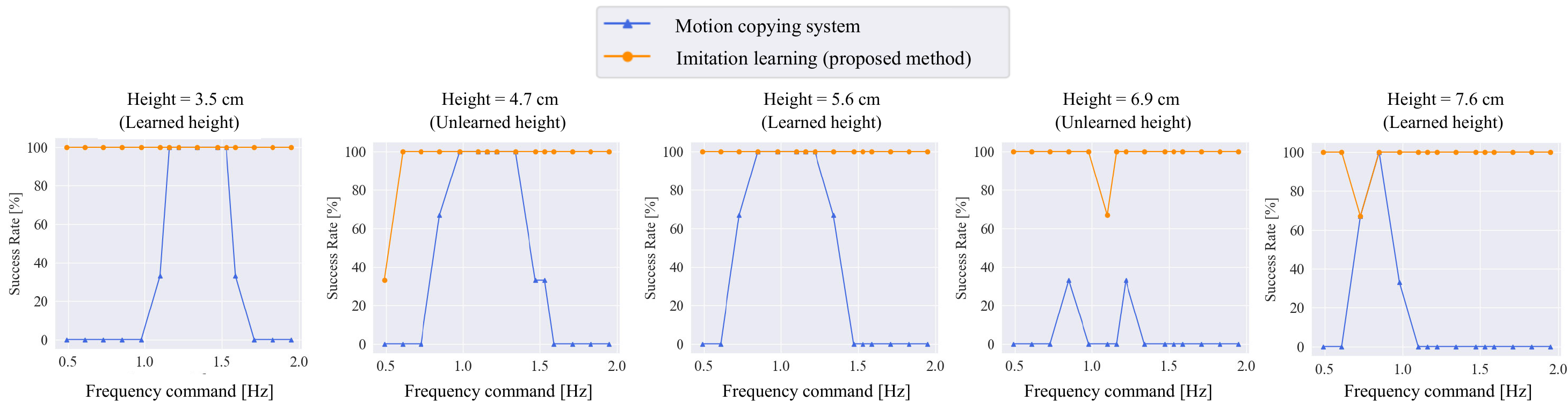}
  \caption{Success rate for erasing lines in autonomous operation.}
  \label{success_rate}
\end{figure*}

\begin{figure*}[t]
  \centering 
  \includegraphics[width=16cm]{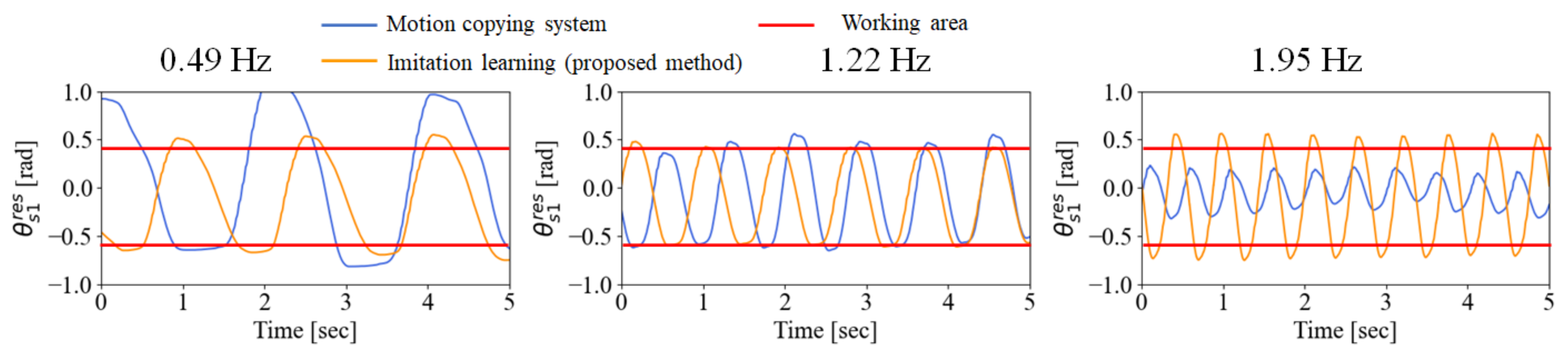}
  \caption{Angular response ($\theta_{s1}^{res} $) in autonomous operation, paper height = 5.6 cm (learned height)}
  \label{exp_theta}
\end{figure*}
\begin{figure*}[t]
  \centering 
  \includegraphics[width=16cm]{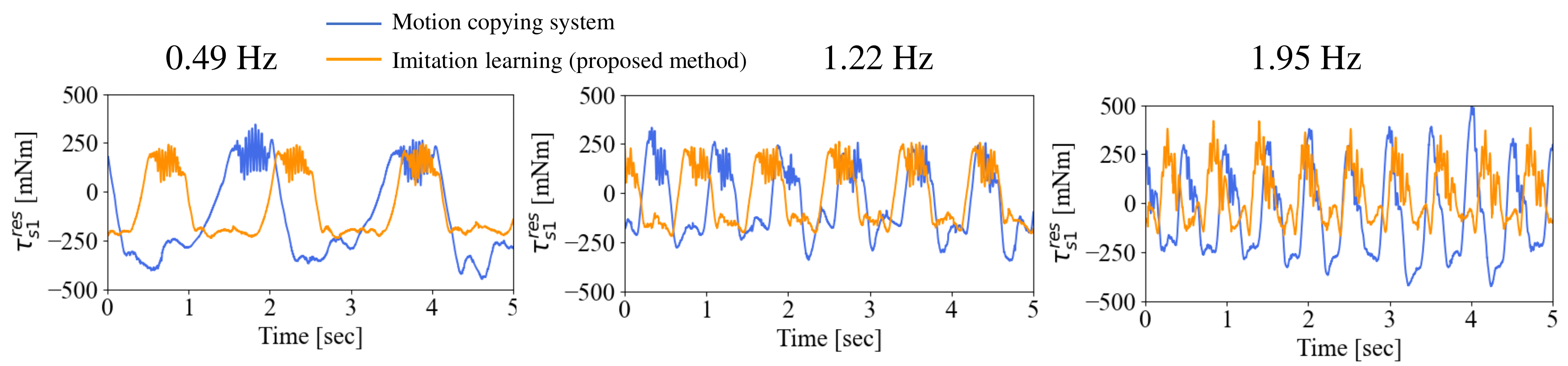}
  \caption{Torque response ($\tau_{s1}^{res} $) in autonomous operation, paper height = 5.6 cm (learned height)}
  \label{exp_tau1}
\end{figure*}
\begin{figure*}[t]
  \centering 
  \includegraphics[width=16cm]{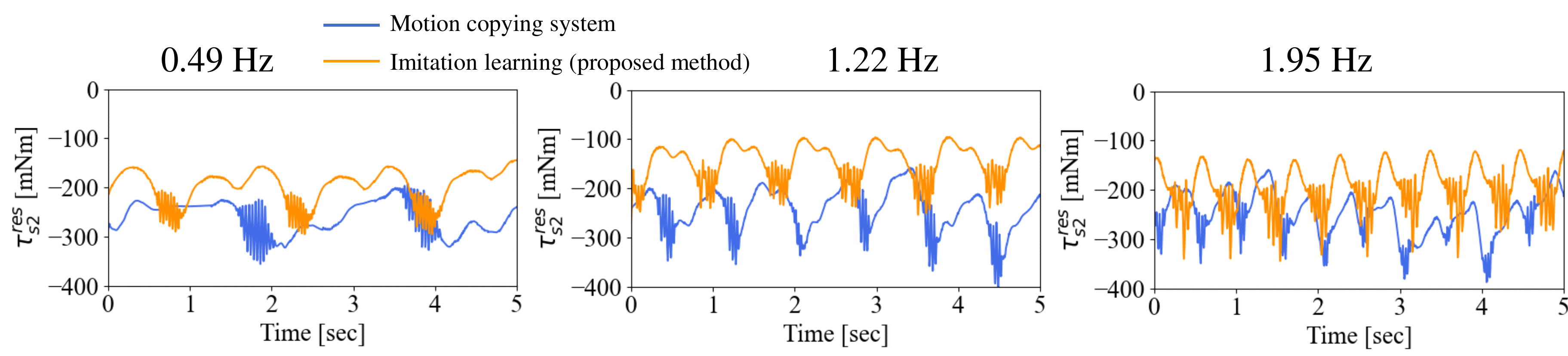}
  \caption{Torque response ($\tau_{s2}^{res} $) in autonomous operation, paper height = 5.6 cm (learned height)}
  \label{exp_tau2}
\end{figure*}
\begin{figure*}[t]
  \centering 
  \includegraphics[width=16cm]{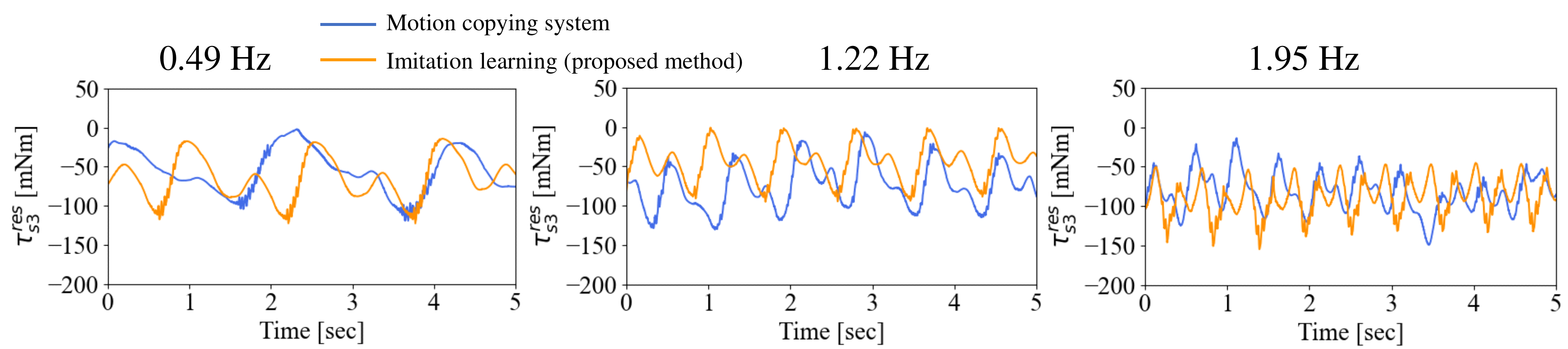}
  \caption{Torque response ($\tau_{s3}^{res} $) in autonomous operation, paper height = 5.6 cm (learned height)}
  \label{exp_tau3}
\end{figure*}

\begin{figure}[t]
  \centering 
  \includegraphics[width=8cm]{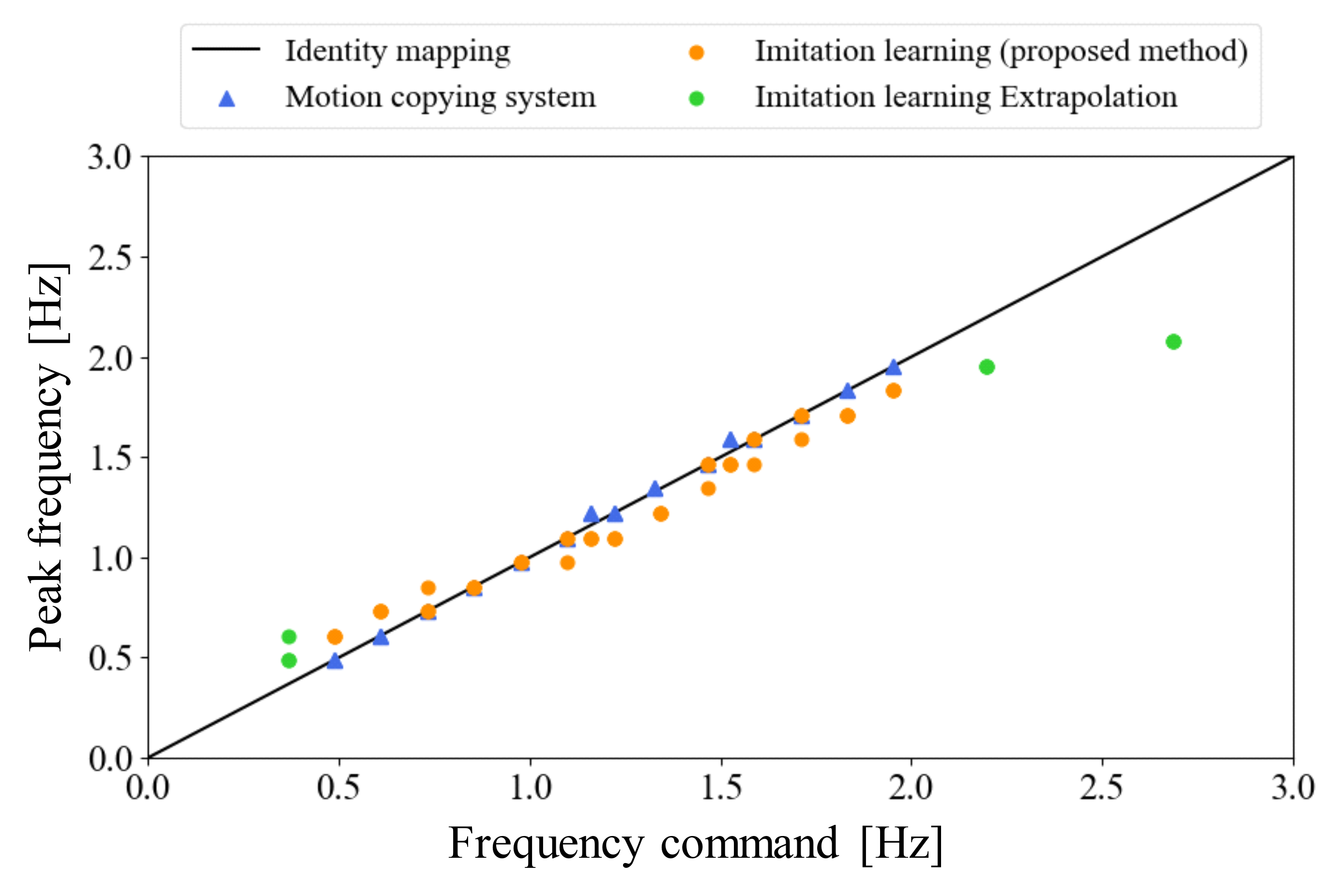}
  \caption{Frequency command and actual frequency at 5.6 cm height in autonomous operation}
  \label{frequency}
\end{figure}

First, nine and 16 convoluted channels were compared for the implementation of the CNN.
The variable-speed range for the 16 channels was wider than that for the nine channels. 
Hence, the proposed method was implemented using 16 channels.
Fig.~\ref{loss} shows the training loss versus epochs for the 16-channel network.
We then selected the model of the 1500 epochs because the loss almost settled down.
The learning time was approximately 40 min. 

\subsection{Results and Discussion}
The success rate of the operation was then evaluated.
Fig.~\ref{area} shows the working area.
Given that this robot was not equipped with a camera, it was not possible to completely erase the entire area.
In contrast, it is easy to erase the entire area by combining the proposed method with conventional methods using a camera. 
However, when several methods are combined, it is difficult to evaluate the effectiveness of the proposed method.
Therefore, we investigated whether we could erase the arc-shaped area through which the end-effector of the robot passed. 
When the robot erased more than 90 \% of the area inside the red lines, this was defined as successful, and it was inferred by a human.
Note that because it is a cyclic motion, it can gradually erase all the lines as time goes by if it could keep operating with stable contact. 

Evaluation was conducted using 15 frequency commands: 0.49, 0.61, 0.73, 0.85, 0.98, 1.10, 1.16, 1.22, 1.34, 1.47, 1.53, 1.59, 1.71, 1.83, and 1.95 Hz, for five heights of 3.5, 4.7, 5.6, 6.9, and 7.6 cm from the surface of the desk.
Three trials were conducted for each condition, for a total of 225 trials (15 [frequencies] $\times$ 5 [heights] $\times$ 3 [trials]).
It should be noted that height information from the desk surface was not given to the robot.
Given that the robot was not equipped with a camera, it needed to adapt to the perturbation of the height using only the angle, angular velocity, and torque information.
The experiments can be viewed using the following link to a video: (https://youtu.be/2XjbauSGu0s).

Fig.~\ref{success_rate} shows the success rate for each height.
The blue lines show the success rates of the motion copying system, whereas the orange lines represent the rates of the proposed method.
As shown in the figure, the motion copying system performs its task under limited frequencies and heights, whereas the proposed method can adapt to variations in both the speed and height.
The success rate was the same or higher than that of the motion-copying system under all conditions.
Particularly, given that the motion copying system does not have an adaptation mechanism against a height perturbation, it was significantly less effective at heights of 6.6 and 7.6 cm.
Figs.~\ref{exp_theta}-\ref{exp_tau3} show the angular responses of $\theta_{s1}^{res}$ and the torque responses of $\tau_{s1}^{res}$, $\tau_{s2}^{res}$, and $\tau_{s3}^{res}$ for a height of 5.6 cm.
The blue lines represent the responses of the motion-copying system, whereas the orange lines show the responses of the proposed method.
The red lines indicate the working area.
In the motion copying system, within a high-speed range, the amplitude of $\theta_{s1}^{res}$ was too small to meet the conditions shown in Fig.~\ref{area}.
In the low-speed range, the amplitude was too large to remain within the desk.
In contrast, in the proposed method, the angular response was almost constant in amplitude and independent of frequency, indicating that it could operate properly within the working range.
However, the amplitudes of the torque varied adaptively for different frequencies in the proposed method.
These figures clearly demonstrate that the proposed method could achieve almost the same angular trajectory regardless of the frequency, whereas the motion copying system exhibited a strong dependency on the frequency; simultaneously, the proposed method could give adaptive force command values with dependencies on the frequency.
Thus, the proposed method can effectively handle frequency-dependent physical phenomena, such as inertial force and friction.
It is also worth noting that the reason why the response of the proposed method and the motion copying system differed at 1.22Hz is that the proposed method generated trajectories in real time, then the motion changed according to subtle changes in the environments and the states of the robot. 
For example, a typical example is that the contact force changed according to the change of the shape of the eraser between the data collection and autonomous operation.
The overall success rate of the proposed method was 98.2 \%.

Fig.~\ref{frequency} shows the reproducibility of the frequency at a height of 5.6 cm.
The horizontal axis shows the frequency command, whereas the vertical axis shows the peak frequency measured using the FFT.
Given that the proposed method was 100 \% successful, all the peak frequencies of $\theta_{s1}^{res}$ are plotted.
Moreover, four additional experiments were conducted to further evaluate the extrapolation performance of the proposed method.
In contrast, given that the conventional method had few successful samples, the behaviors that did not meet the conditions in Fig.~\ref{area} are plotted.
The blue, orange, and green plots show the peak frequencies of the motion copying system, the proposed method, and the proposed method applied during the additional experiment, respectively.
The solid line indicates the identity mapping. When the plots are along the line, the reproducibility of the frequency is ideal.
In the motion copying system, the operating frequency was adjusted by the designer, and the reproducibility of the operating frequency was consequently high.
However, the proposed method could also operate at the command frequency, although there were more variations compared to the motion copying system.
When extrapolation was far from the training data, the reproducibility was reduced, although the peak frequency tended to increase with an increase in the frequency command.
It is worth noting that the operation at 2.08 Hz was achieved using a 2.69 Hz command, indicating that the operation was faster than the fastest training data at 1.83 Hz.
Thus, the proposed method could not only change the operating frequency, but it could also perform the task faster than a human.
Additionally, the control bandwidth of the robot was approximately 2.3 Hz, and it was very difficult to operate at frequencies beyond the control bandwidth.
It is also worth noting that the proposed method can achieve behaviors close to such a limit even when using imitation learning, which has no explicit dynamic model. 
To the best of the authors’ knowledge, there is no other imitation learning that can operate at a frequency almost at the limit of the control bandwidth.

In general imitation learning, the objective is to reproduce the teacher data, thus, the evaluation is mostly based on the success rate of the task, and whether a specific physical quantity can be controlled is not the main issue. However, the proposed method not only imitates the motions but also has a frequency command, thus, the actual frequency can be compared with the frequency command during autonomous operation. Therefore, by labeling the actual frequency as a frequency command and storing it with the motion data, the autonomous motion data can be regarded as new teacher data. If online learning, such as Bayesian optimization, is applied to the new teacher data, the accuracy of motion generation can be further improved iteratively.
In addition, although it is not the subject of this paper, it is expected that generalization performance against modeling errors of the robot's parameters can also be treated as the perturbation of environments.

\subsection{Untrained Object Demonstration}
\begin{figure*}[t]
  \centering 
  \includegraphics[width=15cm]{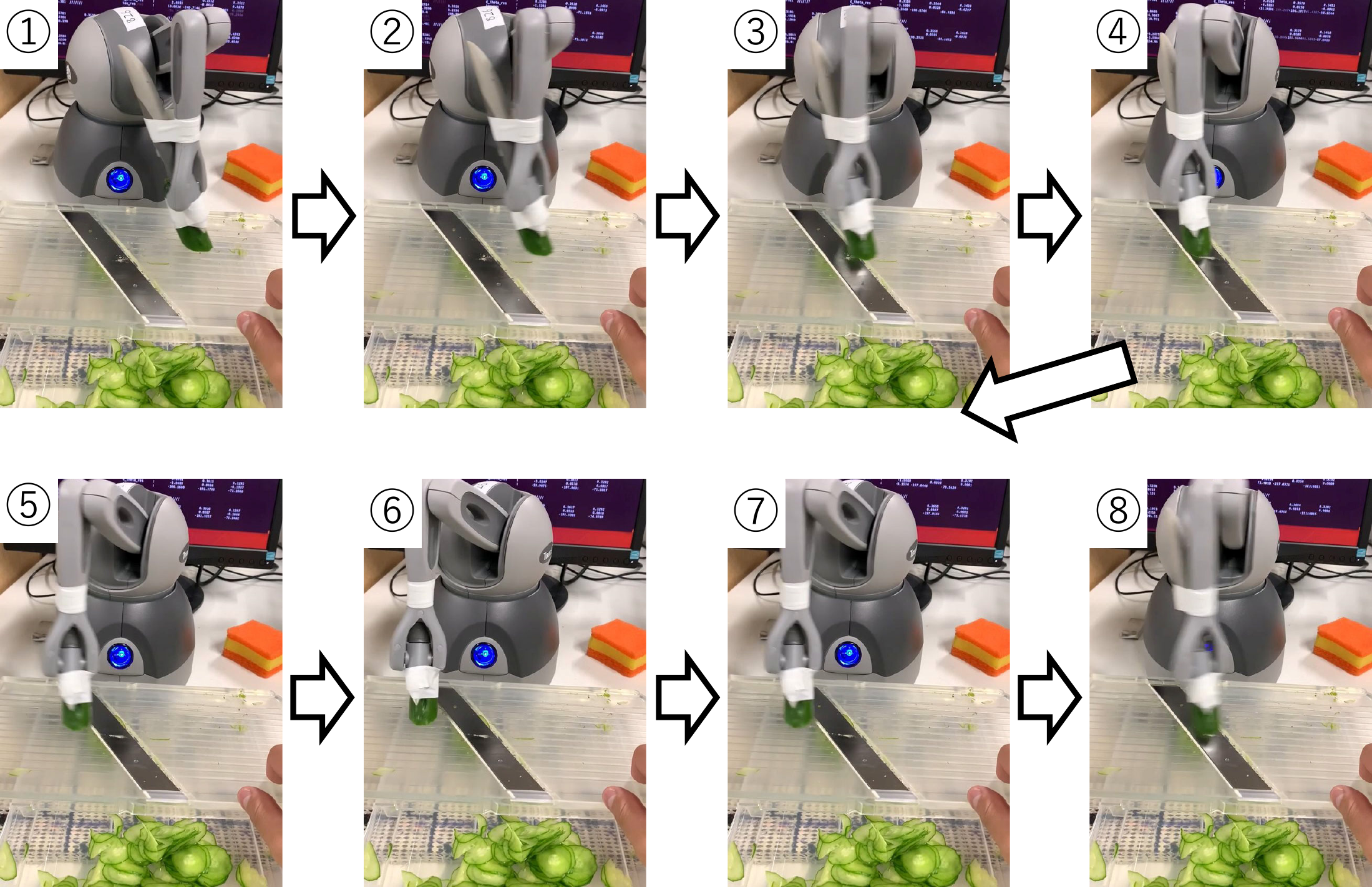}
  \caption{Snapshot of slicing cucumber task}
  \label{snapshot}
\end{figure*}

\begin{figure}[t]
  \centering 
  \includegraphics[width=8cm]{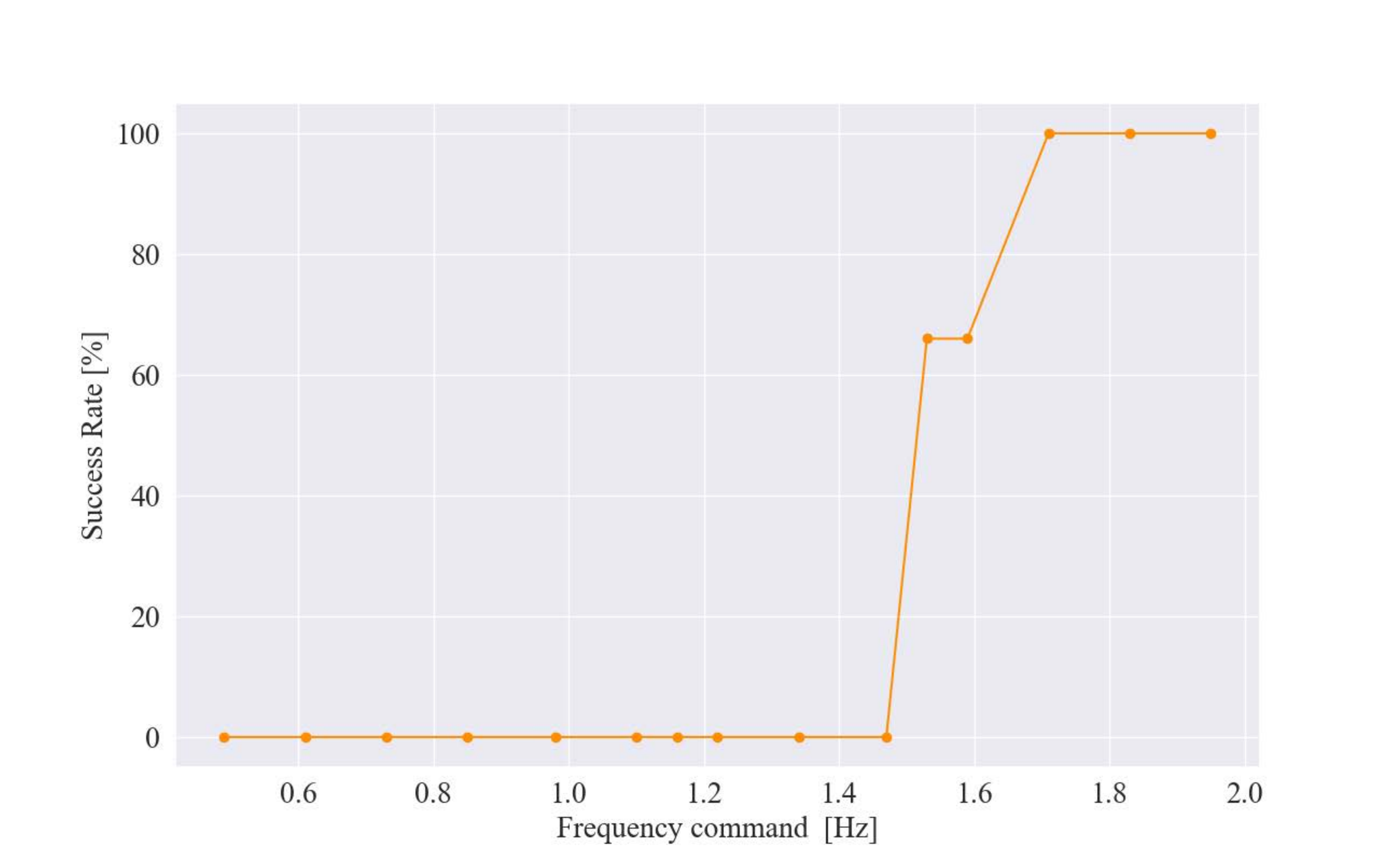}
  \caption{Success rate of slicing cucumber task}
  \label{success_cucumber}
\end{figure}

To further demonstrate the generalization performance of the proposed method, a task of slicing a cucumber was conducted using the same NN model and controller.
No parameter was changed from the previous experiments.
The height of the surface of the slicer was 5.6 cm from the desk surface.
A cucumber was fixed at the tip of the robot, and the cucumber was sliced using a slicer.
Fig.~\ref{snapshot} shows a snapshot of the experiment.
As shown in the figure, the robot continued to slice the cucumber.
The success rates versus frequency commands are shown in Fig.~\ref{success_cucumber}.
The evaluation was conducted using 15 frequency commands: 0.49, 0.61, 0.73, 0.85, 0.98, 1.10, 1.16, 1.22, 1.34, 1.47, 1.53, 1.59, 1.71, 1.83, and 1.95 Hz, and three trials were performed for each command.
This result shows that fast motions were required to slice the cucumber because it was rigid and required a lot of force for slicing.
In other words, the robot had to move quickly and exploit its own inertial force to slice them. 
In conventional force control theories, the inertial forces of robots are often treated as disturbances.
However, using the proposed method, it was possible to realize motions that exploited inertial forces.
Particularly, it is very effective to use the inertial force of a robot with a small maximum output torque, such as the robot used in this experiment.
Additionally, to make the most effective use of the inertial force, a high-frequency motion is required for high acceleration.
It would have been difficult to slice a cucumber without using our method, which can operate the robot to the limit of the control bandwidth.

\section{Conclusion} \label{5:conclusion}
In this study, we proposed a method for generating variable-speed motion while adapting to perturbations in the environment.
Given that there is a nonlinear relationship between the operating speed and frictional or inertial forces, the operating force changes with the work speed. 
Therefore, we confirmed that a variable-speed operation could not be achieved using simple interpolation and extrapolation. 
To solve this problem, we proposed a method to imitate human motion using four-channel bilateral control, a CNN, and an LSTM.
Based on the experimental results, it was determined that the motion varied with the interpolation of the operating speed of the training data as well as the high speed of the extrapolation.
Furthermore, the proposed method can complete a given task faster than a human operator and achieves operations close to the control bandwidth.
The high generalization performance of the proposed method was confirmed by the fact that it was able to acquire the motion of slicing cucumbers from the teacher data of erasing lines with an eraser.
In particular, the success rate of the task was highly dependent on the operating frequency because the cucumber slicing required the explicit use of inertia forces.
The fact that we were able to achieve this operation shows the remarkable characteristics of the proposed method in being able to operate at high speed.
In addition, imitation learning has been evaluated only for generalization performance over space in conventional methods, but this study has made it possible to discuss generalization performance over time.
The approach of analyzing the behavior of the system according to the operating frequency is compatible with the design of ordinary control systems, and the development of a control system design theory that integrates imitation learning and control system design theory is expected.

Note that if the operation is periodic, such that the frequency can be measured in FFT, then this method should be applicable. 
In the future, we will demonstrate the practicality of the proposed method for other periodic motions.
In addition, our another future goal is to improve the reproducibility of the frequency during extrapolation.

\begin{IEEEbiography}[{\includegraphics[width=1in,height=1.25in,clip,keepaspectratio]{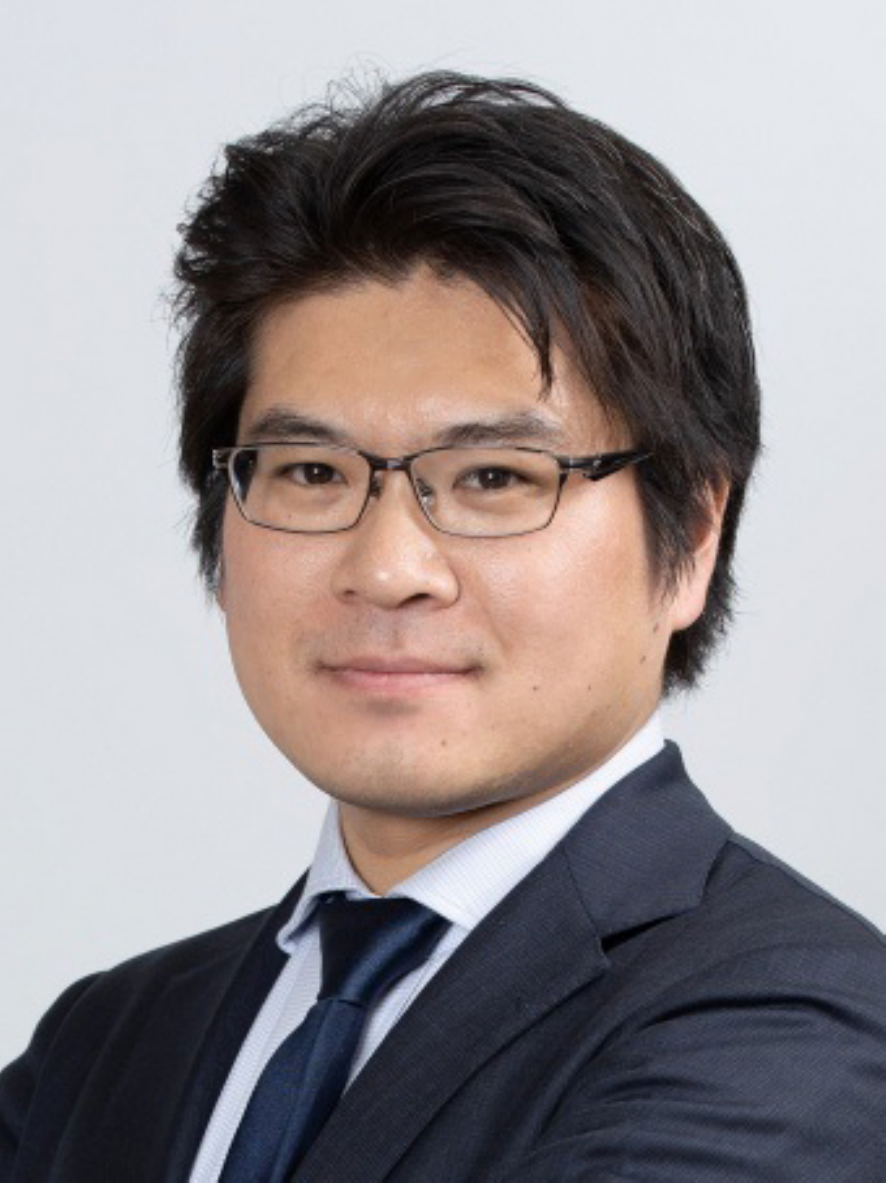}}]{Sho Sakaino}
Sho Sakaino
received the B.E. degree in system design engineering and the M.E.
and Ph.D. degrees in integrated design engineering from Keio University, Yokohama, Japan, in 2006, 2008, and 2011, respectively.
He was an assistant professor at Saitama University from 2011 to 2019.
Since 2019, he has been an associate professor at University of Tsukuba.
His research interests include
mechatronics, motion control, robotics, and haptics. He received the IEEJ Industry Application Society Distinguished Transaction Paper Award in
2011 and 2020. He also received the RSJ Advanced Robotics Excellent Paper Award in 2020.
\end{IEEEbiography}
\begin{IEEEbiography}[{\includegraphics[width=1in,height=1.25in,clip,keepaspectratio]{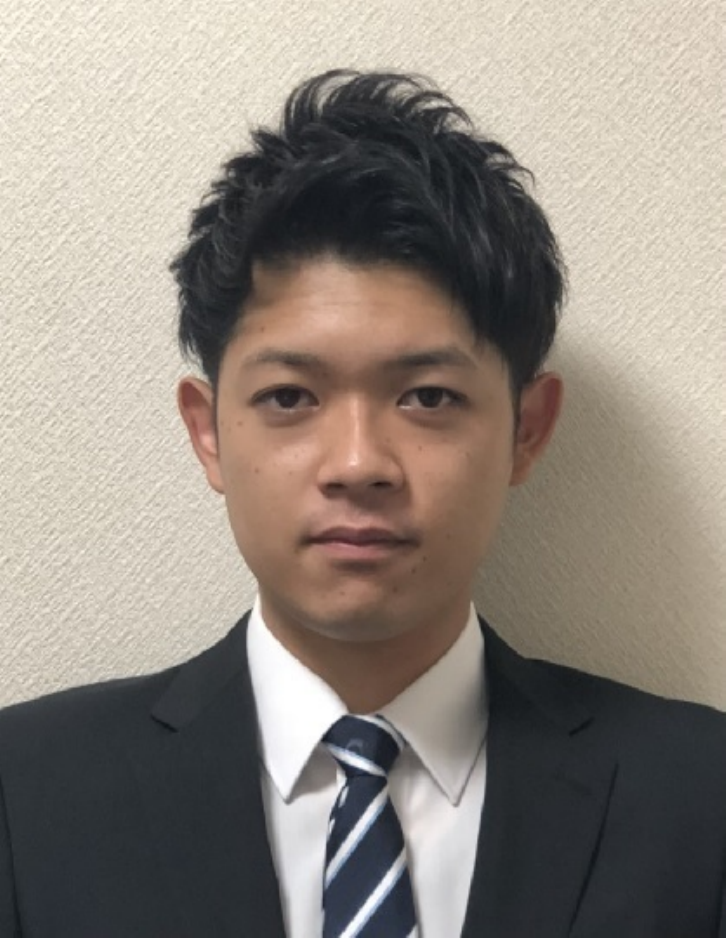}}]{Kazuki Fujimoto}
received the B.E. and M.E. degrees in electrical and
electronic system engineering from Saitama University, Saitama, Japan, in
2018 and 2020, respectively.
\end{IEEEbiography}
\begin{IEEEbiography}[{\includegraphics[width=1in,height=1.25in,clip,keepaspectratio]{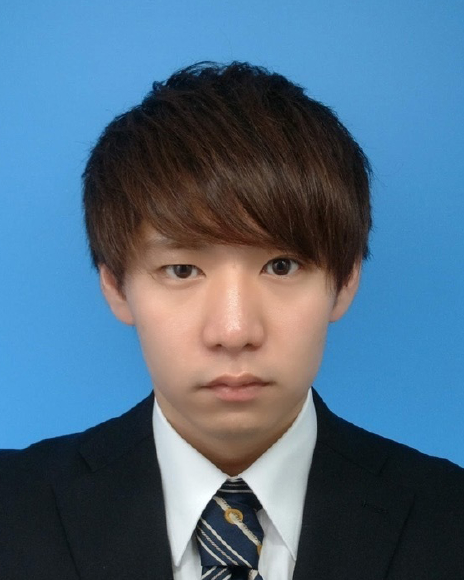}}]{Yuki Saigusa}
received the B.E. degree in electrical and
electronic system engineering from Saitama University, Saitama, Japan, in
2020.
He is currently working toward M.E. degrees in the Graduate School of Science and Technology, and degree programs in intelligent and mechanical interaction systems at University of Tsukuba, Japan.
\end{IEEEbiography}
\begin{IEEEbiography}[{\includegraphics[width=1in,height=1.25in,clip,keepaspectratio]{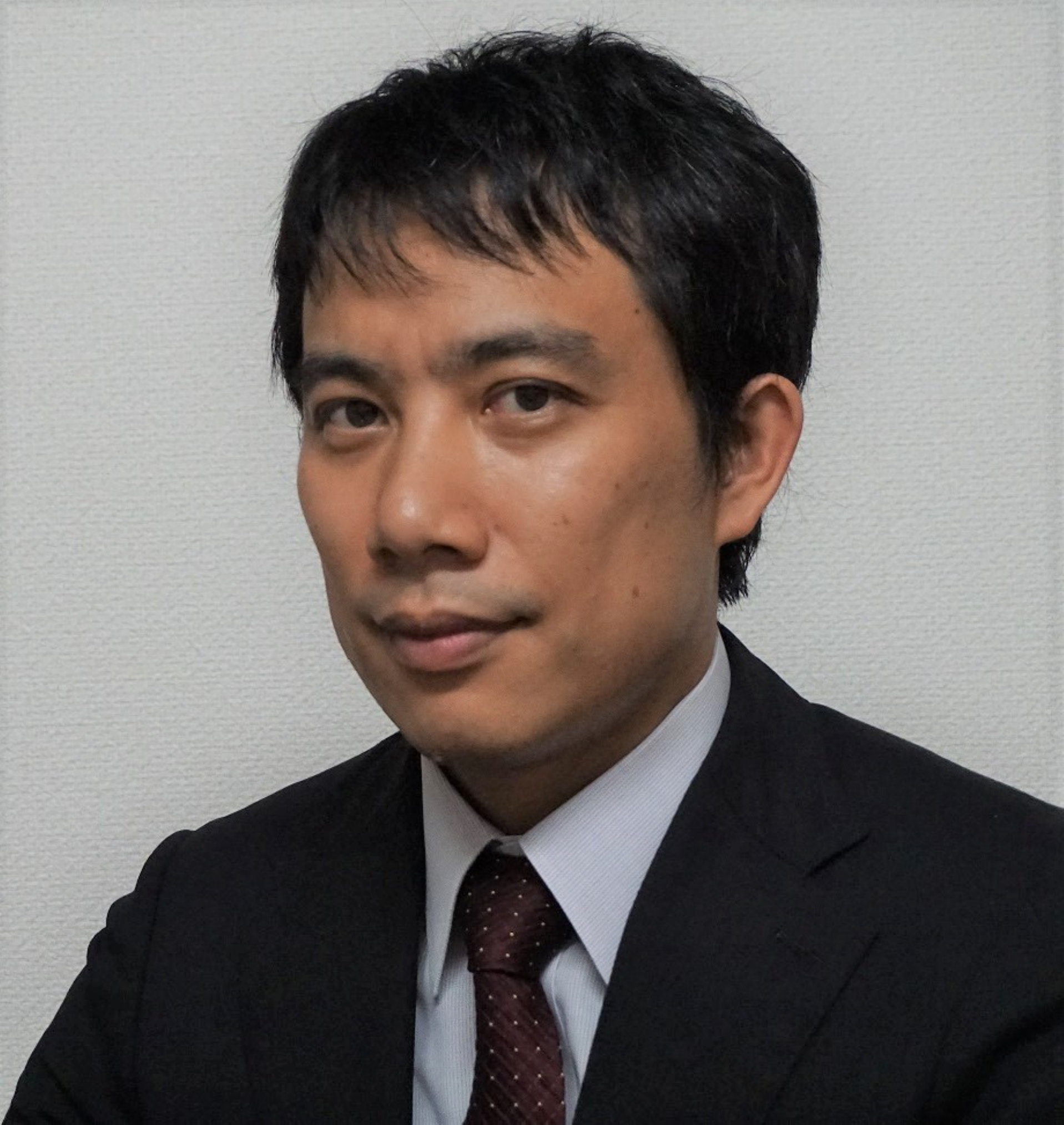}}]{Toshiaki Tsuji}
received the B.E. degree in system design engineering and the
M.E. and Ph.D. degrees in integrated design engineering from Keio University,
Yokohama, Japan, in 2001, 2003, and 2006, respectively. He was a Research
Associate in the Department of Mechanical Engineering, Tokyo University
of Science, from 2006 to 2007. He is currently an Associate Professor in the
Department of Electrical and Electronic Systems, Saitama University, Saitama,
Japan. His research interests include motion control, haptics, and rehabilitation
robots. Dr. Tsuji received the FANUC FA and Robot Foundation Original Paper
Award in 2007 and 2008.
He also received the RSJ Advanced Robotics Excellent Paper Award and the IEEJ Industry Application Society Distinguished Transaction Paper Award in 2020.
\end{IEEEbiography}

\EOD
\end{document}